\pdfoutput=1
\documentclass[11pt]{article}

\usepackage[]{ACL}

\usepackage{times}
\usepackage{tcolorbox}
\usepackage{latexsym}
\usepackage{graphicx} 
\usepackage[T1]{fontenc}
\usepackage[utf8]{inputenc}
\usepackage{subfigure}
\usepackage{subcaption}
\usepackage{microtype}
\usepackage{enumitem}
\usepackage{inconsolata}
\usepackage{amsfonts}
\usepackage{arydshln}
\usepackage{subfigure}
\usepackage{multirow}
\usepackage{float}
\usepackage{CJKutf8}
\usepackage{amsmath}
\usepackage{stmaryrd}

\usepackage[T1]{fontenc}
\usepackage[utf8]{inputenc}

\usepackage{microtype}

\usepackage{inconsolata}

\title{How Far can 100 Samples Go? Unlocking Zero-Shot Translation\\ with Tiny Multi-Parallel Data}

\author{
    Di Wu \quad Shaomu Tan \quad Yan Meng \quad David Stap \quad Christof Monz \\
    Language Technology Lab\\
    University of Amsterdam\\
    \texttt{\{d.wu, s.tan, y.meng, d.stap, c.monz\}@uva.nl}
}

\begin{document}
\maketitle

\begin{abstract}
Zero-shot translation aims to translate between language pairs not seen during training in Multilingual Machine Translation (MMT) and is largely considered an open problem.
A common, albeit resource-consuming, solution is to add as many related translation directions as possible to the training corpus. 
In this paper, we show that for an English-centric model, surprisingly large zero-shot improvements can be achieved by simply fine-tuning with a very small amount of multi-parallel data.
For example, on the EC30 dataset,  we obtain up to +21.7 ChrF non-English overall improvements (870 directions) by using only 100 multi-parallel samples while preserving English-centric translation quality. 
When investigating the size effect of fine-tuning data and its transfer capabilities, we found that already a small, randomly sampled set of fine-tuning directions is sufficient to achieve comparable improvements.
The resulting non-English performance is close to the complete translation upper bound. 
Even in a minimal setting---fine-tuning with only one single sample---the well-known off-target issue is almost completely resolved, explaining parts--but not all---of the observed improvements in translation quality.\footnote{\url{https://github.com/research-anonymous/MultiParallelFinetuning4MMT}}
% Even fine-tuning with only one single sample almost completely resolves the well-known off-target issue, explaining some--but not all---of the observed gains in translation quality.\footnote{\url{https://github.com/research-anonymous/MultiParallelFinetuning4MMT}}
\end{abstract}

\section{Introduction}

% Mainstream status quo
The zero-shot capability shown by Multilingual Machine Translation (MMT)~\cite{johnson2017google} is of considerable significance, particularly in the context of translating between low-resource or distant language pairs. However, even for systems trained on large-scale data, the zero-shot performance is still far from sufficient~\citep{tan2023towards}, especially when scaling up the number of involved languages. 
Substantial efforts~\citep{zhang-etal-2020-improving,pan-etal-2021-contrastive,gu-feng-2022-improving,mao-etal-2023-exploring} have been dedicated to improving the zero-shot capabilities of models trained on readily available, predominantly English-centric corpora.

\begin{figure}[t]
    \centering
    \includegraphics[width=0.40\textwidth]{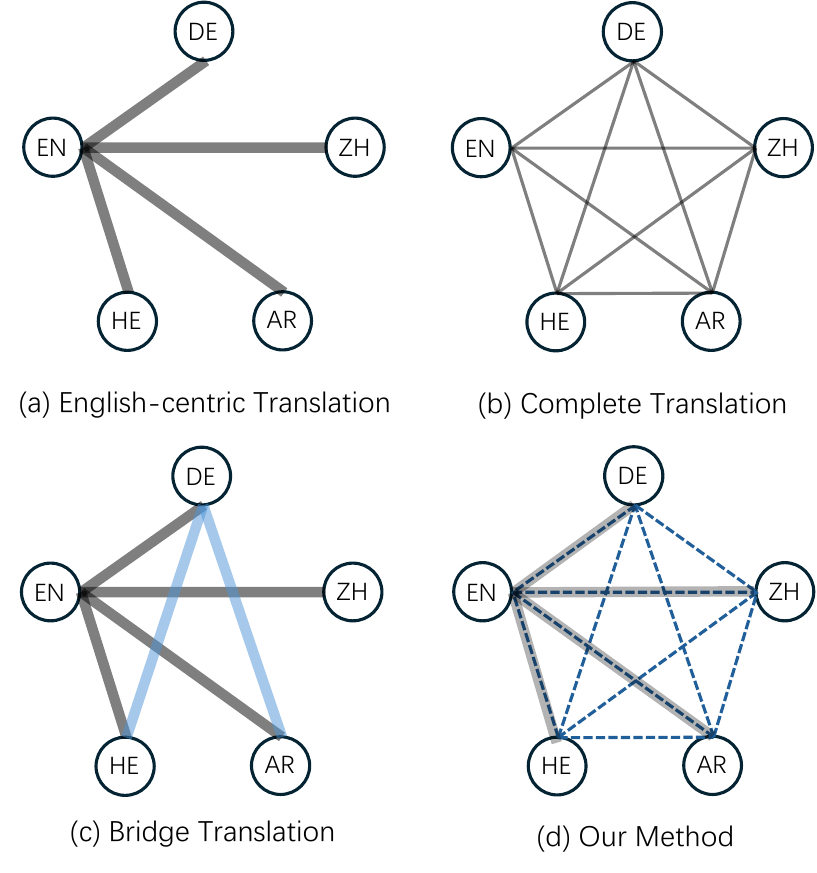}
    \caption{(a) English-centric training data is normally readily available but can only cover a few real-world directions, while (b) complete translation~\citep{freitag-firat-2020-complete} aims to cover all but suffers from the small data scale. (c) Mining partial non-English data as the \textcolor[RGB]{117, 194, 230}{bridge} languages shows promising zero-shot improvements but is also resource-consuming when scaling up. (d) We show that substantial overall improvements can be achieved by fine-tuning an English-centric model with tiny extra multi-parallel data, which is readily available, like NTREX~\citep{federmann-etal-2022-ntrex}.}
    \label{fig:fig0}
\end{figure}

% Complete MMT: focus on mining, not building
To fully cover translation directions, \citet{freitag-firat-2020-complete} propose to mine multi-parallel (multi-way aligned) examples to extend the training set from English-centric to a complete multilingual one as shown in Figure~\ref{fig:fig0}-(b). 
Non-English translation quality in this setting indeed increases substantially. 
However, such a setting is far away from real-world practice when scaling up.
As shown in \citet{freitag-firat-2020-complete}, to solely extend the training set from bilingual aligned to all languages (6-way) involved in their case, the amount of available data drops from 123M to 10K, which is insufficient.

To reflect translation needs worldwide, \citet{fan2021beyond} build and open-source a training dataset covering 100 languages through industry-scale mining. 
In addition to English-centric data, supervised data for thousands of bridge language pairs is mined and included, organized based on language families. The MMT model trained on the resulting data, M2M100, exhibits clear non-English improvement in many non-English directions. 
This work drives a simple but also resource-consuming solution for real-world demand: mining as much training data as possible to bridge non-English language pairs, at least on the language family level.

% This paper
In this paper, we take a step back and again look at the readily available English-centric model. 
% While earlier work has focused on adding translation directions to the training data we are interested in investigating whether it is more effective to include many additional directions during fine-tuning. 
%
We empirically show that the corresponding zero-shot ability can be easily unlocked via fine-tuning an English-centric model with a tiny amount of multi-parallel data, which is much simpler and more efficient than the extensive bridge data mining done by earlier work. 
Furthermore, we investigate the size effect of fine-tuning data: 
1) Surprisingly, even when fine-tuning using randomly sampled 10\% of directions, the overall improvements are comparable to that of full-direction fine-tuning. 
2) The improvements brought by very small fine-tuning datasets only slightly lag behind the upper bound (complete translation) while preserving English-centric capabilities, showing great practical potential. 
3) Even with just one single multi-parallel sample for fine-tuning, the well-known off-target problem \citep{zhang-etal-2020-improving,yang-etal-2021-improving-multilingual,sennrich2023mitigating}, is easily addressed, reducing the off-target rate from 51.8\% to 1.9\%. 
However, not all improvements in translation quality can be solely attributed to lower off-target rates as we also see clear improvements in cases where translations are already in the correct target language.

Due to the high efficiency and practicality, we encourage the community to consider fine-tuning with tiny readily available multi-parallel data, like NTREX~\citep{federmann-etal-2022-ntrex}, as a strong baseline for zero-shot translation. 
%Also, we call on the community to build more comprehensive multi-parallel datasets to cover real-world demands.

\section{Related Work}
The zero-shot translation capability of MMT is associated with multilingualism, following the hypothesis of universal representation or interlingua.
\citet{arivazhagan2019missing} view zero-shot translation as a domain adaptation problem~\citep{ben2006analysis} in MMT, and apply auxiliary losses to explicitly incentivize the model to learn and use domain- (language-) invariant representation.
\citet{liu-etal-2021-improving-zero} attribute the low quality of zero-shot MT to the positional correspondence to input tokens, which hinders modeling language-agnostic representation.
\citet{pan-etal-2021-contrastive} use a contrastive loss to close the representation gap between different languages. 
Some other approaches aim to harness the capabilities of pretrained multilingual models for zero-shot translation. \citet{chen-etal-2022-towards} employ multilingual pretrained encoders to extend bilingual translation to many-to-one translation, relying on the encoder's language-agnostic representation. 
Recently, some work has focused on leveraging pre-trained large language models for multilingual translation~\citep{zhang-etal-2023-machine,moslem2023adaptive}. 
Despite the inclusion of the so-called ``emergent abilities''~\citep{wei2022emergent} triggered by zero-shot prompting, we categorize these works as following a similar line.

This paper focuses on a data-centric approach for comprehensively improving zero-shot performance. 
We empirically show that a well-trained English-centric model can be easily boosted for overall zero-shot capability via fine-tuning with minimal multi-parallel data, even if only covering a small set of translation directions (10\%). 
This allows us to leverage multi-parallel data which is hard to obtain in large quantities, leading to a highly efficient and practical means for overall zero-shot translation.
We note that \citet{maillard-etal-2023-small} also show that integrating small high-quality data (6K samples) into the training corpus can have a big impact on low-resource translation systems, especially when combined with back translation~\citep{sennrich-etal-2016-improving}. However, we argue that the reasons for the effectiveness differ: 1) as shown in Section~\ref{how_close_to_the_upper_bound}, the substantial enhancements persist when using data built from the training set, meaning that the influence from domain or quality level is eliminated, and 2) our method can work with extremely minimal fine-tuning data (100 or even a single sample).

\section{Experiments}\label{experiments}

In this paper, we propose a simple approach that leverages small amounts of multi-parallel data to fine-tune an English-centric model for a large number of directions. The fine-tuning data is constructed from small, readily available multi-parallel datasets, like NTREX~\citep{federmann-etal-2022-ntrex}. We refer to such a process as ``fine-tuning with multi-parallel data'' or ``multi-parallel fine-tuning''.

% In this paper, we propose to straightforwardly fine-tune an English-centric model in various directions using bitexts constructed from a very small, readily available multi-parallel dataset, like NTREX~\citep{federmann-etal-2022-ntrex}. We refer to such a process as "fine-tuning with multi-parallel data" or "multi-parallel fine-tuning".

% \subsection{Multi-Parallel Fine-Tuning}
\subsection{Fine-Tuning Data Construction}
% In this paper, we propose a simple approach that leverages small amounts of multi-parallel data to fine-tune an English-centric model for a large number of translation directions. The fine-tuning data is constructed from a very small, readily available multi-parallel dataset, like NTREX~\citep{federmann-etal-2022-ntrex}. 

% In this paper, we propose to straightforwardly fine-tune an English-centric model in various directions using bitexts constructed from a very small, readily available multi-parallel dataset, like NTREX~\citep{federmann-etal-2022-ntrex}. We refer to such a process as ``fine-tuning with multi-parallel data'' or ``multi-parallel fine-tuning''.

Given a multi-parallel dataset comprising $N$ distinct languages, each with $K$ samples, we can generate pairwise data in all $N \times (N-1)$ possible directions. 
Note that acquiring large quantities of multi-parallel data poses challenges due to many professional human translators being involved. 
However, horizontally expanding a readily available multi-parallel dataset to include one more language is straightforward. 
It simply requires annotating $K$ additional samples for the new language based on the current dataset, with $2 \times N$ new translation directions indirectly covered. 
% In this paper, we refer to ``fine-tuning with multi-parallel data'' or ``multi-parallel fine-tuning'' as leveraging the resulting pairwise data to fine-tuning an English-centric model for a large number of directions.
% Also, our fine-tuning is efficient: most experiments in this paper are finished within 1 GPU hour.

\subsection{Datasets} \label{datasets}
\paragraph{NTREX-128.} NTREX\footnote{\url{https://github.com/MicrosoftTranslator/NTREX}}~\citep{federmann-etal-2022-ntrex} is initially proposed as an evaluation dataset, expanding multilingual testing for translation from English into 128 target languages, which consists of 1997 samples per language and mainly focus on the News domain. 
Given the multi-parallel organization of NTREX data, we can easily build arbitrary pairwise data across 128 languages. 
In this paper, we leverage NTREX to create our fine-tuning datasets and conduct experiments to highlight the big impact of such a tiny amount of data.

\paragraph{Europarl-8.} Europarl\footnote{\url{https://www.statmt.org/europarl}}~\citep{koehn-2005-europarl} consists of 20 English-centric language pairs from the proceedings of the European Parliament, with sizes ranging from 399K to 2M. 
A characteristic of Europarl is that part of the samples are multi-way aligned. 
In this paper, we select the most resource-rich 8 languages, i.e., EN, DA, DE, ES, FI, FR, IT, and NL, to mine a fully multi-parallel dataset named Europarl-8 via aligning multi-way sentences with the same English part. 
This results in about 1.2M fully multi-parallel data instances, where each sentence has 7 counterparts in other languages. 

\paragraph{EC30.} To ensure a more diverse and inclusive large-scale evaluation, we follow \citet{tan2023towards,wu-monz-2023-beyond} and use the EC30 dataset, which is built from WMT~\citep{bojar-etal-2017-findings} and OPUS~\citep{tiedemann-2012-parallel} corpora. 
EC30 comprises 61 million English-centric bilingual sentences for training, encompassing 30 non-English languages with diverse resource levels (High: 5M, Medium: 1M, Low: 100K). Each resource group includes languages from 5 families with multiple writing systems.

\paragraph{Evaluation Benchmark.} For all of the experiments in this paper, we evaluate translations via the Flores-101 benchmark~\citep{goyal-etal-2022-flores}.
Flores comprises 3001 sentences sourced from English Wikipedia, which covers a variety of topics and domains and is translated into 101 languages by professional translators. 
We use \emph{dev} and \emph{devtest} as the validation and test dataset, consisting of 997 and 1012 samples, respectively.
All results are evaluated on three widely used metrics, namely, ChrF++~\citep{popovic2017chrf++}, SacreBLEU~\citep{post-2018-call}\footnote{nrefs:1|case:mixed|eff:no|tok:13a|smooth:exp|version:2.3.1}, and COMET~\citep{rei-etal-2020-comet}, to demonstrate the consistency of improvements across a broad spectrum of evaluation metrics.
A more detailed description of the datasets is provided in Appendix~\ref{appendix_detailed_dataset_description}.

% Table-EC30: Zero-shot (ChrF)
\begin{table*}[t]
\centering
\scalebox{0.85}{
\begin{tabular}{l|l|ccccccccc|c}
\hline\hline
& Model &H-H & H-M & H-L & M-H & M-M & M-L & L-H & L-M & L-L & AVG \\ 
\hline
\multirow{5}{*}{\rotatebox{90}{Two-Tag}} & Baseline &11.0 & 14.8 & 10.6 & 11.3 & 14.9 & 10.5 & 13.7 & 17.4 & 10.9 & 12.8 \\
& Boost-100 &37.9 & 39.1 & 30.6 & 38.0 & 38.8 & 30.5 & 33.9 & 34.8 & 26.8 & 34.5 \\
& Boost-All &38.6 & 39.9 & 32.0 & 38.7 & 39.5 & 31.8 & 34.9 & 35.9 & 28.4 & 35.5 \\ 
\cdashline{2-12}
& $\Delta$-100  &+26.9 & +24.3 & +20.0 & +26.7 & +23.9 & +20.0 & +20.2 & +17.4 & +15.9 & +21.7 \\
& $\Delta$-All  &+27.6 & +25.1 & +21.4 & +27.4 & +24.6 & +21.3 & +21.2 & +18.5 & +17.5 & +22.7 \\
\hline
\multirow{5}{*}{\rotatebox{90}{One-Tag}} & Baseline &28.0 & 30.4 & 20.0 & 27.9 & 29.5 & 19.7 & 24.0 & 26.0 & 16.4 & 24.7 \\
& Boost-100 &36.6 & 37.4 & 29.0 & 36.5 & 36.9 & 28.8 & 32.1 & 32.7 & 24.8 & 32.8 \\
& Boost-All &37.2 & 38.2 & 30.6 & 37.3 & 37.8 & 30.5 & 33.4 & 34.0 & 26.9 & 34.0 \\ 
\cdashline{2-12}
& $\Delta$-100  &+8.6 & +7.0 & +9.0 & +8.6 & +7.4 & +9.1 & +8.1 & +6.7 & +8.4 & +8.1 \\
& $\Delta$-All  &+9.2 & +7.8 & +10.6 & +9.4 & +8.3 & +10.8 & +9.4 & +8.0 & +10.5 & +9.3 \\
\hline
\hline
\end{tabular}}
\caption{\label{table-EC30-zero-shot-ChrF} Zero-shot performance (ChrF) on the EC30 dataset (870 directions, 61M sentence pairs), grouped by \textbf{H}igh-, \textbf{M}edium, and \textbf{L}ow-resource, respectively. $\Delta$-100 and $\Delta$-All mean the corresponding performance changes compared to the baselines. Results in SacreBLEU and COMET are provided in Table~\ref{table-appendix-EC30-SacreBLEU} and Table~\ref{table-appendix-EC30-COMET}, respectively.}
\end{table*}

\subsection{Experimental Setup}
\subsubsection{Training Setting}
For experiments on the EC30 dataset, we use Transformer-Big with 16 attention heads, 1,024 embedding dimensions, and 4,096 feedforward dimensions. 
For Europarl-8, we use a smaller backbone, as the training data is smaller, where a standard 6-layer encoder, 6-layer decoder transformer model is applied with 4 attention heads, 512 embedding dimensions, and 1,024 feedforward dimensions. 
In total, 447M and 64M  training parameters are involved for the two models. 
More detailed training settings are provided in Appendix~\ref{appendix_training_setting}.

\subsubsection{Fine-Tuning Setting} 
We use full-parameter fine-tuning and keep our setup as simple as possible to highlight generalizability. 
We reset all running statuses, including optimizer, lr scheduler, and data loaders. 
Also, the fine-tuning parameters are aligned with those in the training period, except for the experiments in Section~\ref{analysis}, where we set batch accumulation as 1 as extremely small fine-tuning data is used.

% Table-EC30: En-centric (ChrF)
\begin{table*}[t]
\centering
\scalebox{0.85}{
\begin{tabular}{l|l|cc|cc|cc|ccc}
\hline\hline
\multirow{2}{*}{} & \multirow{2}{*}{Model} & \multicolumn{2}{c|}{High} & \multicolumn{2}{c|}{Medium} & \multicolumn{2}{c|}{Low} & \multicolumn{3}{c}{Overall} \\ 
\cline{3-11} 
& & EN-X & X-EN & EN-X & X-EN & EN-X & X-EN & EN-X & X-EN & AVG \\ \hline
\multirow{5}{*}{\rotatebox{90}{Two-Tag}} & Baseline & 52.5 & 57.5 & 53.9 & 56.5 & 42.5 & 49.8 & 49.6 & 54.6 & 52.1 \\
& Boost-100 & 51.9 & 56.6 & 53.6 & 57.0 & 42.6 & 50.2 & 49.4 & 54.6 & 52.0 \\
& Boost-All & 51.9 & 56.0 & 53.6 & 56.6 & 43.2 & 50.9 & 49.6 & 54.5 & 52.0 \\ \cdashline{2-11}
& $\Delta$-100 & -0.6 & -0.9 & -0.3 & +0.5 & +0.1 & +0.4 & -0.2 & 0.0 & -0.1 \\
& $\Delta$-All & -0.6 & -1.5 & -0.3 & +0.1 & +0.7 & +1.1 & 0.0 & -0.1 & -0.1 \\ \hline
\multirow{5}{*}{\rotatebox{90}{One-Tag}} & Baseline & 52.6 & 57.0 & 54.0 & 56.2 & 42.9 & 49.6 & 49.8 & 54.3 & 52.1 \\
& Boost-100 & 52.0 & 56.1 & 53.7 & 56.5 & 43.2 & 49.9 & 49.6 & 54.2 & 51.9 \\
& Boost-All & 51.7 & 55.7 & 53.5 & 56.2 & 43.5 & 50.3 & 49.6 & 54.1 & 51.8 \\ \cdashline{2-11}
& $\Delta$-100 & -0.6 & -0.9 & -0.3 & +0.3 & +0.3 & +0.3 & -0.2 & -0.1 & -0.2 \\
& $\Delta$-All & -0.9 & -1.3 & -0.5 & 0.0 & +0.6 & +0.7 & -0.3 & -0.2 & -0.3 \\ \hline
\end{tabular}}
\caption{\label{table-EC30-EN-centric-ChrF} English-centric performance (ChrF) on the EC30 dataset (60 directions, 61M sentence pairs). EN-X and X-EN denote the average out-of- and into-English translation performance of each resource group, respectively. Results in SacreBLEU and COMET are provided in Table~\ref{table-EC30-EN-centric-SacreBLEU} and Table~\ref{table-EC30-EN-centric-COMET}, respectively.}
\end{table*}

\begin{figure*}[t]
  \begin{minipage}{0.5\textwidth}
    \centering
    \subfigure[Increasing the number of directions (\textsf{D})]{\includegraphics[width=8cm]{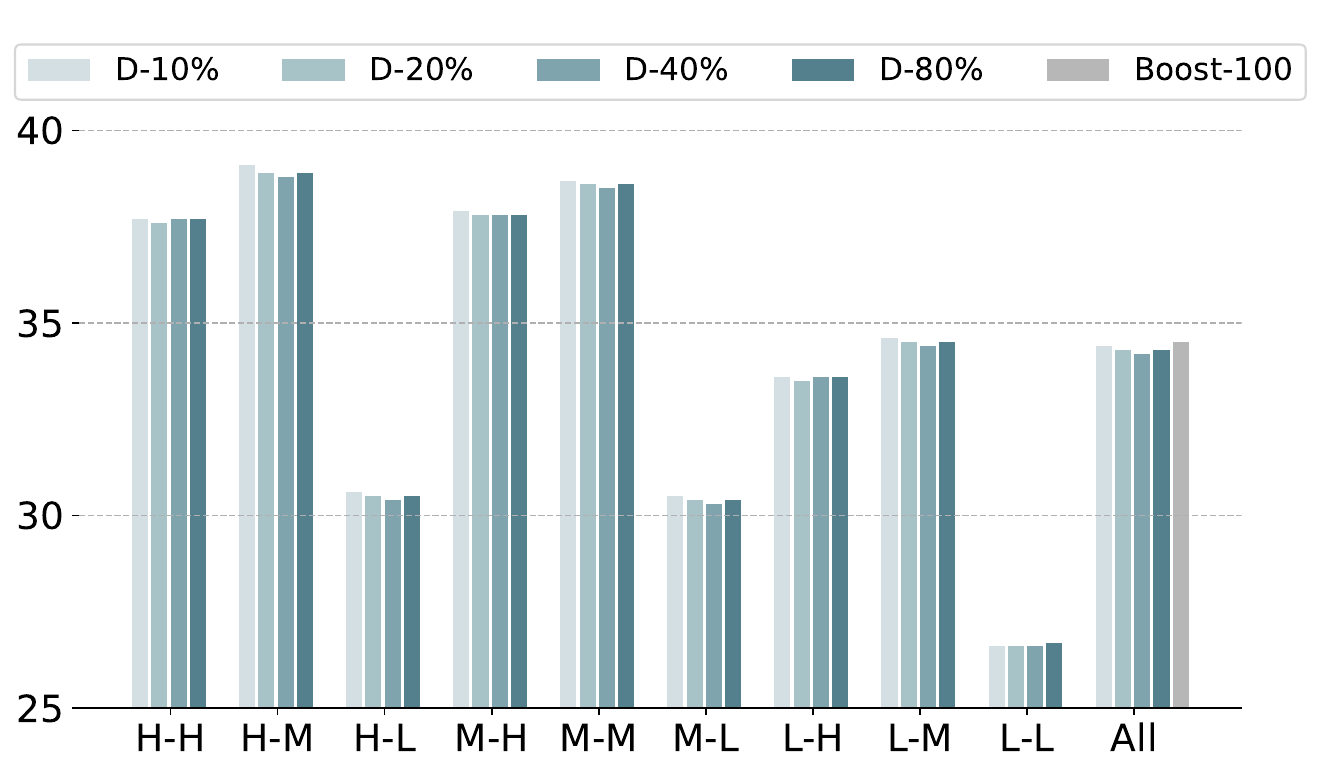}}
  \end{minipage}%
  \begin{minipage}{0.5\textwidth}
    \centering
    \subfigure[Increasing the number of samples (\textsf{S})]
    {\includegraphics[width=8cm]{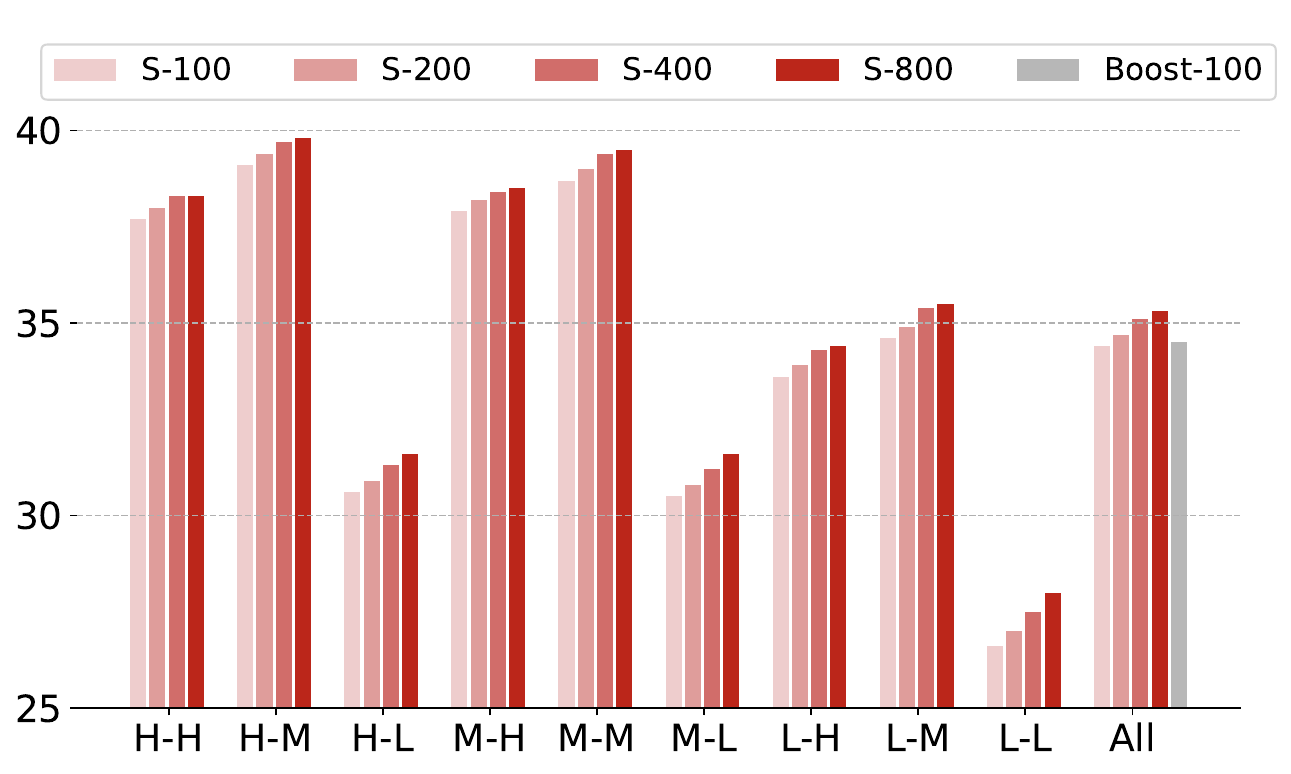}}
  \end{minipage}
  \caption{Zero-shot performance (ChrF) on EC30 for each scaling step, grouped by \textbf{H}igh-, \textbf{M}edium, and \textbf{L}ow-resource, respectively. (a) When we randomly selected \{10\%, 20\%, 40\%, 80\%\} of fine-tuning directions, overall zero-shot performance nearly stayed unchanged. However, (b) when we fixed 10\% of directions and increased the fine-tuning samples from 100 to 800, consistent improvements can be observed for all resource groups.}
  \label{fig:fig5}
\end{figure*}

\subsection{Large-Scale Experiments on EC30}\label{EC30_main}
In this section, we show how far a tiny amount of multi-parallel data can improve the zero-shot capability of an already well-trained large-scale English-centric MMT system. 
We conduct experiments on EC30, involving 30 English-centric and 870 zero-shot directions. 
We build fine-tuning data based on NTREX to cover all except Occitan-related directions---Occitan is not included in NTREX---as described in Section~\ref{datasets}.
It is noteworthy that MMT systems typically use two language tag strategies: 1) the one-tag strategy, i.e., adding the target language IDs to the encoder input, which is shown by \citet{wu-etal-2021-language} to be more effective for zero-shot translation, or 2) the two-tag strategy, i.e., adding source and target language IDs to the encoder and decoder input, respectively, which is often applied to recent large-scale MMT systems~\citep{fan2021beyond,pan-etal-2021-contrastive,costa2022no}. 
To show comprehensive results, we conducted experiments in both settings.

Table~\ref{table-EC30-zero-shot-ChrF} shows the zero-shot performance across 9 resource groups. 
Boost-All means using all of the 1,997 multi-parallel samples from NTREX to construct pair-wise fine-tuning data for all directions (including English-centric ones), while Boost-100 means only using 100 randomly sampled samples, instead of 1,997, to construct the fine-tuning data.

We find that 1) fine-tuning with tiny data leads to very strong overall improvements for both tagging strategies, with up to +9.3 and +22.7 average ChrF point gains, respectively. 
2) The zero-shot capability of the two-tag baseline lags behind the one-tag baseline, in line with \citet{wu-etal-2021-language}. 
However, after fine-tuning with multi-parallel data, the overall performance in the two-tag setting consistently outperforms the one-tag setting for each group, yielding an average margin of +1.5 ChrF (35.5 v.s., 34.0). 
Consistent improvements also hold for other metrics, see Appendix~\ref{results_on_EC30}.

In Table~\ref{table-EC30-EN-centric-ChrF}, we show the impact of fine-tuning on English-centric directions: The trade-off effect mainly occurs in the high-resource group. However, the influence on the medium- and low-resource groups is negligible or even positive, especially for the low-resource part, resulting in nearly unchanged overall English-centric performance. 
For instance, fine-tuning with 100 multi-parallel samples on the two-tag model yields +21.7 ChrF zero-shot gains, with negligible drops in averaged English-centric performance (-0.1 ChrF). 
In Table~\ref{table-appendix-EC30-weak-ChrF}, we show that 854 out of 870 zero-shot directions get strong boosts (more than 10.0 ChrF).

It is noteworthy that fine-tuning with just 100 samples achieves comparable improvements to using the entire NTREX dataset (+21.7 v.s.\ +22.7 in Table~\ref{table-EC30-zero-shot-ChrF}), even though the latter's size is 20 times larger.
This diminishing effectiveness naturally leads us to ask (i) whether more fine-tuning data or more fine-tuning directions is important and (ii) how close can our method come to the upper-bound improvements.

% It is noteworthy that fine-tuning with just 100 samples achieves comparable improvements to using the entire NTREX dataset (+21.7 v.s. +22.7 in Table~\ref{table-EC30-zero-shot-ChrF}), although the latter's size is 20 times larger. 
% This diminishing effectiveness naturally leads us to ask (i) more fine-tuning data or more fine-tuning directions are important and (ii) how much our method can approximate the upper bound if we further increase the amount of multi-parallel data.

We answer both questions in Section~\ref{more_data_or_more_directions} and \ref{how_close_to_the_upper_bound}, respectively.
If not specified, we employ the two-tag strategy in subsequent experiments because of its higher zero-shot and English-centric performance after fine-tuning.

\subsection{More Data or More Directions?}\label{more_data_or_more_directions}
In Section~\ref{EC30_main}, we showed that fine-tuning an English-centric model with a small amount of bitext derived from NTREX (covering all directions) yields substantial zero-shot improvements.
A natural assumption is that the improvement in each direction is triggered by the corresponding directional data. 
In this section, we investigate whether this is true, i.e., what will happen if we only cover a subset of translation directions during fine-tuning.

We conduct experiments on the same English-centric model trained on EC30, see Section~\ref{EC30_main}, and control the scale of the fine-tuning data in the following two settings: (a) We conducted a random sampling of 100 multi-parallel NTREX sentences to construct pairwise data to cover all 870 directions\footnote{These directions include English-centric ones while excluding Occitan-related ones as it is not available in NTREX.}. 
Then, we randomly sampled \{10\%, 20\%, 40\%, 80\%\} directions for fine-tuning. 
(b) We fixed the 10\% of directions as mentioned in (a) and conducted a random sampling of \{100, 200, 400, 800\} multi-parallel NTREX instances to construct the fine-tuning set for the corresponding directions.

% \begin{enumerate}[label=(\alph*)]
%     \item We conducted a random sampling of 100 multi-parallel NTREX sentences to construct pairwise data to cover all 870 directions\footnote{These directions include English-centric ones while excluding Occitan-related ones as it is not available in NTREX.}. Then, we randomly sampled \{10\%, 20\%, 40\%, 80\%\} directions for fine-tuning.
%     \item We fixed the 10\% of directions as mentioned in (a) and conducted a random sampling of \{100, 200, 400, 800\} multi-parallel NTREX instances to construct the fine-tuning set for the corresponding directions.
% \end{enumerate}

Note that the bitext size in settings (a) and (b) for each scaling step is kept identical, e.g., to facilitate a fair comparison with the setting of fine-tuning with 100 multi-parallel samples in 80\% directions, we also consider fine-tuning in 10\% directions with 800 multi-parallel samples.

\begin{figure*}[tb]
    \centering
    \includegraphics[width=0.95\textwidth]{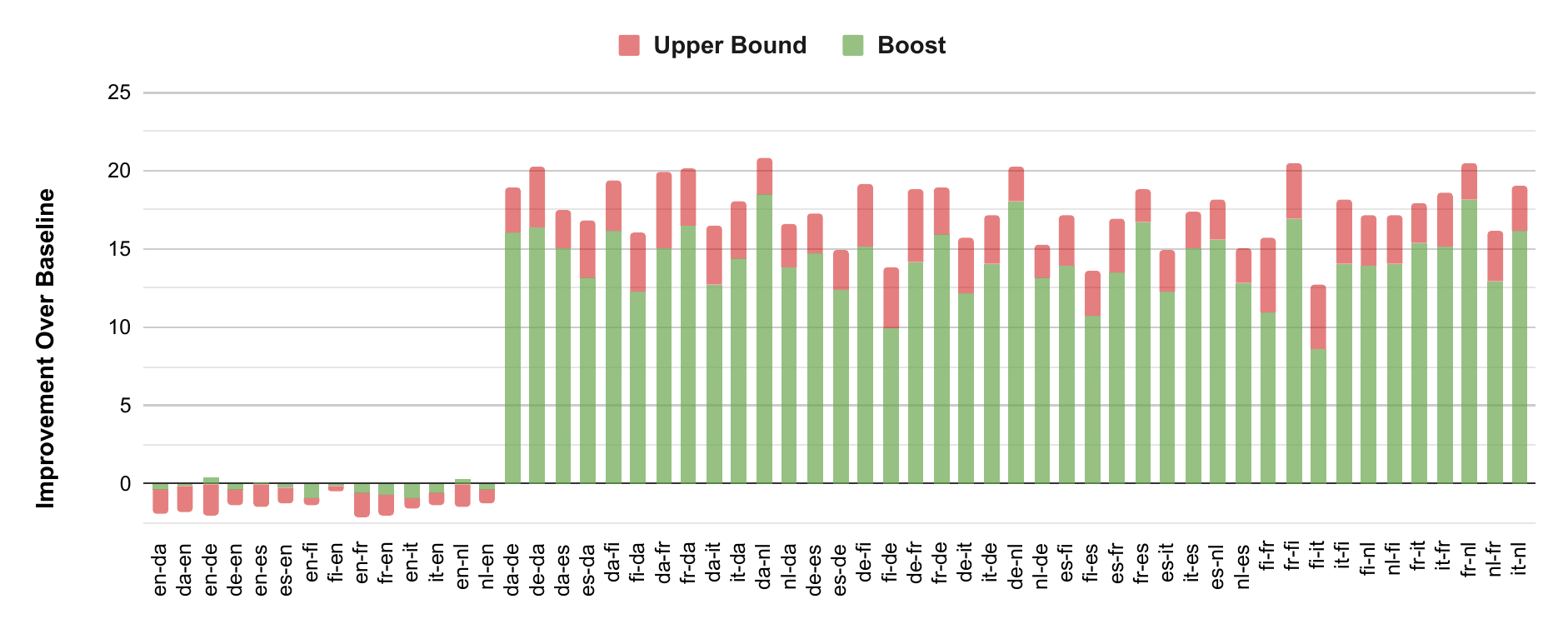}
    \caption{ChrF improvements of the \textcolor[RGB]{236, 132, 129}{upper bound} and \textcolor[RGB]{159, 192, 137}{boosted} models over the English-centric baseline on the Europarl-8 dataset. It is clear that the overall non-English capability of the \textcolor[RGB]{159, 192, 137}{boosted} model is close to the \textcolor[RGB]{236, 132, 129}{upper bound} (complete translation), meanwhile, it also holds the performance in English-centric directions.}
    \label{fig:fig2}
\end{figure*}

We show all of the corresponding fine-tuning results in Figure~\ref{fig:fig5}. 
Surprisingly, when fixing the size of the multi-parallel samples to 100 and then increasing the fine-tuning directions from 10\% to 80\%, no improvement is observed for any resource group (Figure~\ref{fig:fig5}-a). 
Fine-tuning in randomly sampled 10\% directions using 100 samples achieves comparable overall results to fine-tuning in all directions (Boost-100).
However, when we fix the directions to 10\% and increase the multi-parallel sample size from 100 to 800, consistent improvements for all groups can be observed (Figure~\ref{fig:fig5}-b).
This shows that the overall improvements are not sensitive to the number of directions, at least when the directions extend to a certain scale, like 10\%.

In Appendix~\ref{Limied_Direction_Set}, we further show that when we limited the fine-tuning direction set to fall in a specific family (Germanic), overall improvements just slightly lag behind that of fully fine-tuning, showing surprising cross-lingual transfer ability.

\subsection{How Close to the Upper Bound?}\label{how_close_to_the_upper_bound}
In this section, we show to what extent our fine-tuning method can approximate an upper bound.
Here, we consider the performance of complete translation, i.e., training with fully multi-parallel data, as the "upper bound", since identical scales of non-English bitext cover all of the directions that the English-centric counterparts can not cover.

We conduct experiments on Europarl-8, see Section~\ref{datasets}, where 8-way aligned data are available. 
Both English-centric and complete translation models are trained based on it. 
Note that we reuse the former's vocabulary for the latter to ensure a fair comparison. 
Also, we present the results after fine-tuning the English-centric model using full-direction pairwise data constructed from NTREX. 

In Figure~\ref{fig:fig2}, we show that 1) the upper-bound performance, i.e., that of the complete model, surpasses the baseline by a large margin, resulting in +17.4 average ChrF gains for non-English directions. 
2) However, the boosted model's performance closely approaches the upper bound in all non-English directions. 
3) For the 14 English-centric directions, both the upper bound and the boosted models exhibit degradation compared to the baseline, which reveals trade-off effects from English-centric to non-English directions. 
However, the boosted method only slightly degrades for a few English-centric directions (e.g., en-fi and en-it in Figure~\ref{fig:fig2}), whereas the upper bound model drops for most. 
Detailed scores, including those in other metrics, are provided in Appendix~\ref{appendix_how_close}. 
In short, the boosted model achieves strong non-English gains (+14.3 ChrF) with a negligible cost in English-centric directions (-0.3 ChrF).
% In short, the boosted model achieves strong non-English gains (+14.3 ChrF, 42 directions) with a negligible cost in English-centric directions (-0.3 ChrF).

\begin{figure}[tb]
    \centering
    \includegraphics[width=0.46\textwidth]{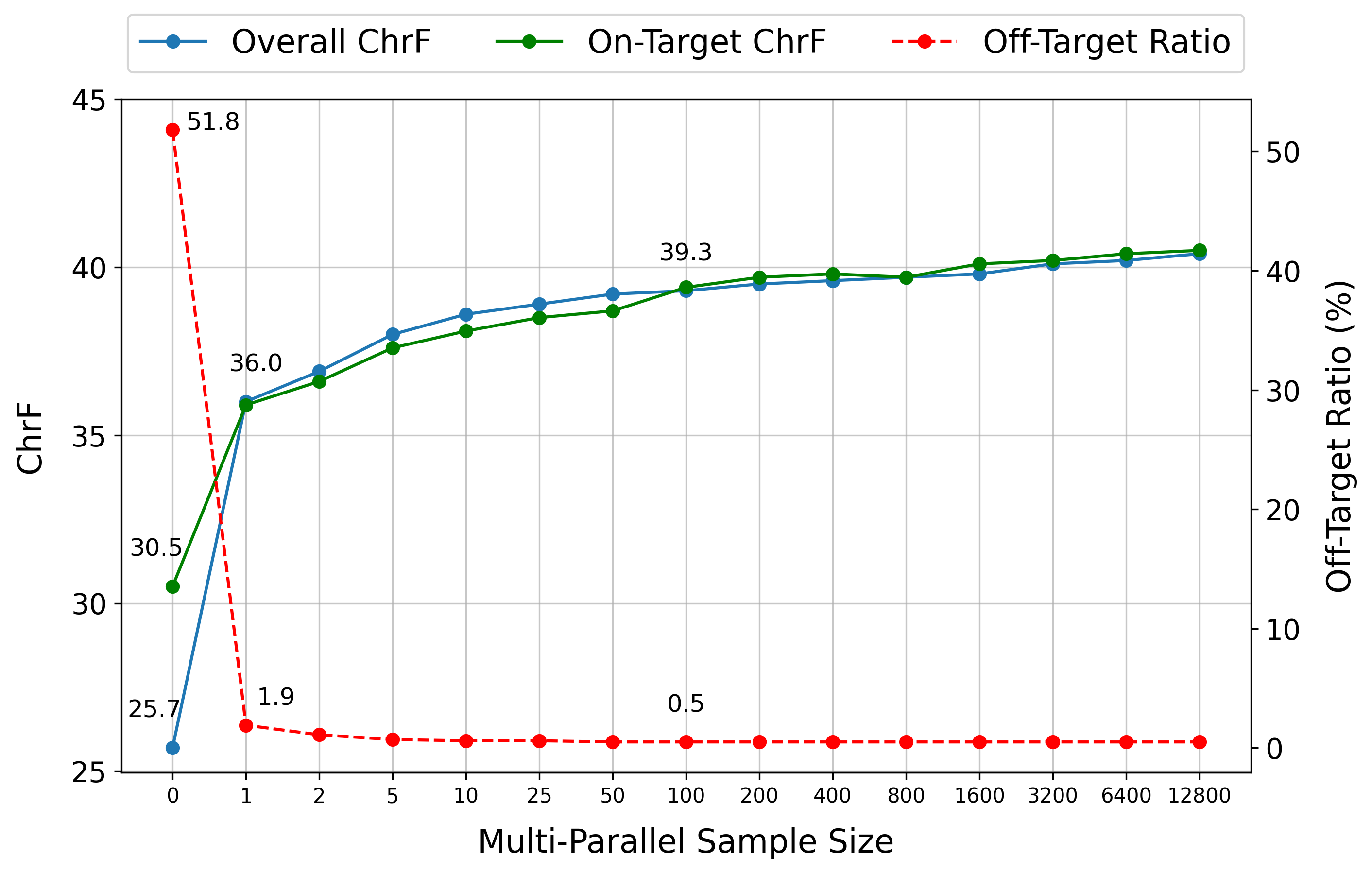}
    \caption{Zero-shot performance and off-target ratio on Europarl-8 at each scaling step. The \textcolor[RGB]{55, 125, 34}{green solid line} denotes the quality improvements of the translation samples that have no off-target issue.}
    \label{fig:fig4}
\end{figure}

\section{Analysis}\label{analysis}
\subsection{Off-Target and Fine-Tuning Data Size}
The off-target problem is often viewed as a primary cause that impairs the zero-shot capability~\citep{zhang-etal-2020-improving,yang-etal-2021-improving-multilingual,sennrich2023mitigating,chen-etal-2023-target}. 
In this section, we delve into the impact of fine-tuning data size on off-target ratios and final performance. 
Moreover, we disentangle the gains of the already on-target translations, showing the extent to which the enhancements are beyond alleviating the off-target issue. 
Note that since Europarl-8 is fully multi-parallel, we can readily build the corresponding full-direction fine-tuning data at different scales.

% \begin{table*}[t]
% \centering
% \scalebox{0.9}{
% \begin{tabular}{l|ccccccccc|c|c}
% \hline
% \hline
% Model & H-H & H-M & H-L & M-H & M-M & M-L & L-H & L-M & L-L & ZS-AVG & EN-AVG \\
% \hline
% Baseline & 11.0 & 14.8 & 10.6 & 11.3 & 14.9 & 10.5 & 13.7 & 17.4 & 10.9 & 12.8 & 52.1\\
% \hdashline
% (a) Multi-Parallel & 37.6 & 38.9 & 30.1 & 36.7 & 37.3 & 29.0 & 32.1 & 32.9 & 24.8 & 33.3 & 50.9\\
% (b) Multi-Directional & 37.6 &\bf 39.1 &\bf 30.3 &\bf 37.1 &\bf 38.0 &\bf 29.7 &\bf 33.0 &\bf 34.1 &\bf 26.0 &\bf 33.9 & 51.5 \\
% \hline
% \hline
% \end{tabular}}
% \caption{\label{table-2} Decoupling multi-parallel and multi-directional fine-tuning on the EC30 dataset. ZS-AVG and EN-AVG denote the average results of the zero-shot and English-centric performance in ChrF, respectively.} 
% \end{table*}

\begin{table*}[t]
\centering
\scalebox{0.9}{
\begin{tabular}{l|ccccccccc|c|c}
\hline
\hline
Model & H-H & H-M & H-L & M-H & M-M & M-L & L-H & L-M & L-L & ZS-AVG & EN-AVG \\
\hline
Baseline & 11.0 & 14.8 & 10.6 & 11.3 & 14.9 & 10.5 & 13.7 & 17.4 & 10.9 & 12.8 & 52.1\\
\hdashline
(a) Multi-Parallel & 37.6 & 38.9 & 30.1 & 36.7 & 37.3 & 29.0 & 32.1 & 32.9 & 24.8 & 33.3 & 50.9\\
(b) Multi-Directional & 37.6 & 39.1 & 30.3 & 37.1 & 38.0 & 29.7 & 33.0 & 34.1 & 26.0 & 33.9 & 51.5 \\
\hline
\hline
\end{tabular}}
\caption{\label{table-2} Decoupling multi-parallel and multi-directional fine-tuning on the EC30 dataset. ZS-AVG and EN-AVG denote the average results of the zero-shot and English-centric performance in ChrF, respectively.} 
\end{table*}

Here, we sample multiple sets of multi-parallel instances from the training set of Europarl-8, ranging from 1 to 12.8K with different seeds 3 times. 
The average results in ChrF after fine-tuning for each scaling step are provided in Figure~\ref{fig:fig4}. 
Also, we report the corresponding off-target ratio evaluated by fastText\footnote{\url{https://github.com/facebookresearch/fastText}}~\citep{joulin-etal-2017-bag} following previous works~\citep{yang-etal-2021-improving-multilingual,costa2022no}.

%\textcolor[RGB]{57, 119, 175}{blue solid line}
%\textcolor[RGB]{55, 125, 34}{green solid line}
The blue solid line shows the overall zero-shot performance at each scaling step, where the starting point (fine-tuning with 0 samples) denotes the performance of the original English-centric model. 
Notably, a high off-target ratio (51.8\%) exists at this point. 
Surprisingly, even fine-tuning with just one multi-parallel sample, very strong overall zero-shot improvements can be obtained (from 25.7 to 36.2 ChrF). 
Meanwhile, the off-target issue is almost completely resolved, dropping from 51.8\% to 1.9\%. 
Increasing from 1 to 100 samples, we can still observe clear zero-shot capability boosting (from 36.2 to 39.3 ChrF), while the off-target change is marginal. 
Further scaling up fine-tuning data from the point of 100 samples shows nearly linear performance gains.

We further disentangle the evaluation set into on-target parts%
\footnote{Note that the on-target sample size varies across directions, with an average of 488 samples per direction.%
} 
for each language direction, where each source sample is already translated into the correct direction when evaluating the English-centric model. 
In Figure~\ref{fig:fig4}, the green solid line denotes the average performance on the on-target part. 
It is easy to see that the improvements brought from fine-tuning with one single sample are still strong, even after isolating the impact of off-target issues. 
As the number of fine-tuning samples increases, on-target improvements closely follow the trend of the overall improvements, further showing that the overall improvements are not only due to resolving the off-target issue.

\subsection{Does Multi-Parallelism Matter?}
% In Section~\ref{experiments}, we showed surprising boosting effects from the additional tiny multi-parallel samples. However, does the semantically equivalent signal matter here? In this section, we delve deeper into the question of whether utilizing multi-parallel data, as opposed to trivial pairwise data, for fine-tuning is crucial for substantial enhancements.

We have shown remarkable boosting effects obtained from adding just a tiny amount of multi-parallel data. 
But, does the data have to be \emph{multi}-parallel? In this section, we explore whether utilizing multi-parallel data, instead of just pairwise data, for fine-tuning is vital for significant enhancements.

To this end, we fine-tune the English-centric model built on EC30 in 5 languages (20 directions). The fine-tuning data is also built from NTREX. However, we control the resulting bitext distribution.
Firstly, we randomly map the 1,997 multi-parallel samples into 10 buckets (roughly 100 samples for each). Then, we construct pairwise data in the following two ways:

\begin{enumerate}[label=(\alph*)]
    \item Multi-Parallel: We constructed pairwise samples in the 20 directions using the multi-parallel data in one randomly picked bucket.
    \item Multi-Directional: For each bucket, we construct fine-tuning samples for a specific language pair only (2 directions), also resulting in the 20 translation directions.
\end{enumerate}

Note that the size of the bitext in settings (a) and (b) are identical. 
In (a), each sentence has semantically equivalent counterparts in all other languages. 
However, in (b), each sentence has only one counterpart, resulting in simple pairwise data.

To cover different language families and resource levels, we choose DE, FR, RU, HE, and AR as these 5 languages. 
In Table~\ref{table-2}, we show the results that fine-tune the EC30-based English-centric model with data in (a) multi-parallel and (b) multi-directional settings, respectively.
Firstly, compared to the baseline model, clear improvements can be observed for zero-shot translation in both settings. 
Meanwhile, in all groups, the performance in setting (a) closely trails but never surpasses that in setting (b). It shows that the boosting effects do not depend on multi-way semantic equivalence, showing simple multi-directional data is sufficient in case fully multi-parallel samples do not exist.

\begin{table}[t]
\centering
\scalebox{0.8}{
\begin{tabular}{l|cc|c|c}
\hline
\hline
Setting & EN-X & X-EN & Zero-Shot & Off-Target (\%) \\
\hline
Baseline & 49.8 & 51.3 & 25.7 & 51.8\\
\hdashline
Numbers & 49.9 & 51.5 & 26.8 & 46.1 \\
Words & 48.3 & 50.7 & 35.8 & 3.6 \\
NTREX & 49.5 & 50.9 & 40.0 & 0.5 \\
\hline
\hline
\end{tabular}}
\caption{\label{table-insights} The results (ChrF) after fine-tuning. "Numbers", "Words", and "NTREX" denote different types of fine-tuning data (see Section~\ref{some_insights}).}
\end{table}

\subsection{The Role of Semantic and Syntactic Information in Fine-Tuning Data}\label{some_insights}
Considering that a small amount of fine-tuning data, e.g., 100 or even one single sample, can still substantially enhance overall zero-shot performance, a related question arises: To what extent do these improvements stem from the intrinsic information inherent in the data itself? 
In this section, we provide some insights into the role that the semantics and syntactic of fine-tuning data play in the unexpected improvements for zero-shot translation. 

We choose the English-centric model trained on Europarl-8 as our baseline (see Section~\ref{how_close_to_the_upper_bound}), and fine-tune it on three datasets as follows:

\paragraph{Number Pairs.} For each direction, we perform uniform sampling of digits (ranging from 1 to 1000) multiple times, concatenating them to a certain length.
Then, it is replicated for both the source and target sides, forming a number translation sample, as shown in Figure~\ref{fig:number_pairs}. 
Given no semantic other than numerical information is contained in this setting, we try to check whether the improvements stem from factors other than the data itself, e.g., the tags.

\paragraph{Word Pairs.} We utilize bilingual dictionaries from MUSE~\footnote{\url{https://github.com/facebookresearch/MUSE}}~\citep{lample2018word} to build word pairs for all of the directions. 
MUSE contains 110 English-centric bilingual dictionaries and all languages of Europarl-8 are included. 
We first select the intersection of English words for the 7 involved English-centric dictionaries. 
Then, we extend them by mapping words paired with the same English words together.\footnote{Specifically, for each one-to-many mapping that exists, we randomly select a one-to-one mapping.}
E.g., given an EN-DE pair \{\textit{bike}, \textit{Fahrrad}\} and an EN-NL pair \{\textit{bike}, \textit{fiets}\}, we can build a new DE-NL word pair \{\textit{Fahrrad}, \textit{fiets}\}. 
Finally, we built 28 dictionaries (16,737 word pairs for each) covering all 56 directions.

\paragraph{Sentence Pairs.} We use 100 randomly selected multi-parallel samples from NTREX to construct pairwise data covering all directions.

To ensure a fair comparison, we maintain similar surface information across the three datasets, such as aligning the number of tokens in the number-pair and word-pair datasets with the English portion in the sentence-pair dataset. 
Table~\ref{table-insights} shows the corresponding fine-tuning results: 
1) Fine-tuning with number pairs results in marginal improvements. 
Conversely, fine-tuning with word pairs leads to noticeable zero-shot improvements (+10.1 ChrF). 
Simultaneously, the off-target ratio also decreases to an acceptable level. 
This means that semantic information, particularly at the lexical level, plays an important role here. 
2) When using sentence-pair data (NTREX) to fine-tune, considerable further improvements compared to word-pair counterparts can be observed, showing that syntactic-level information also matters.

\section{Conclusion}
In this paper, we show that the zero-shot performance of an English-centric MMT model can be easily boosted by a tiny amount of multi-parallel data. On EC30, +21.7 ChrF average gains can be achieved by fine-tuning using 100 samples from NTREX, meanwhile preserving the English-centric performance, see Section~\ref{EC30_main}. More surprisingly, we show that fine-tuning on a small portion (10\%) of directions can achieve comparable improvements to full-direction fine-tuning, see Section~\ref{more_data_or_more_directions}, which are even close to the ideal but impractical upper-bound model, see Section~\ref{how_close_to_the_upper_bound}.

In terms of using language tags, we show that fine-tuning can address the two-tag model's performance degradation in zero-shot directions~\citep{wu-etal-2021-language}. 
Moreover, the final performance substantially surpasses that of the one-tag model across multiple metrics, see Section~\ref{EC30_main}.

We also question earlier findings~\citep{zhang-etal-2020-improving,yang-etal-2021-improving-multilingual,sennrich2023mitigating,chen-etal-2023-target} that consider the off-target issue as a challenging problem for MMT.  This paper shows that the off-target issue can be easily addressed by fine-tuning with tiny (even single) multi-parallel samples. Lastly, we shed some light on the impact of different types of fine-tuning data on the final performance. 

Given the clear advantages of our proposed method, we encourage the community 1) to consider the use of fine-tuning as a strong baseline for zero-shot translation in the future, especially for the two-tag setting, and 2) to construct more comprehensive and high-quality multi-parallel datasets that cover real-world demands.

\section*{Limitations}
Multi-parallel data are normally built in a way that translates the same English data into multiple other languages by professional human translators. Hence, in the resulting non-English fine-tuning data, both the source and target side are translated instead of using the original text. This may exacerbate potential drawbacks in certain directions, such as translations

\section*{Broader Impact}
MMT systems have significant progress recently. However, potential challenges such as mistranslation or off-target issues still exist. Moreover, the fairness problem also arises, e.g., the generation ability is not guaranteed to be fair across languages or demographic features, which may run the risk of reinforcing societal biases, e.g., race bias.

% \section*{Ethics Statement}
% xxx

% \section*{Acknowledgements}
% xxx

\bibliography{anthology,custom}
\bibliographystyle{acl_natbib}

\appendix

\section{Appendix}
\label{sec:appendix}

\subsection{Detailed Dataset Description}\label{appendix_detailed_dataset_description}
\paragraph{EC30.} In Table~\ref{table-appendix-1}, we list the details of the EC40 dataset. We conducted experiments on EC30, a subset of EC40, where we excluded the data of 10 super low-resource languages, resulting in 30 English-centric language pairs with a total of 61M pairwise data. Each resource group consists of languages from 5 families with multiple writing systems.

\subsection{Training Setting}\label{appendix_training_setting}
For all of the English-centric training, the learning rate is 5e-4 with 4,000 warmup steps and a \emph{inverse sqrt} decay schedule. All dropout rates and label smoothing are set to 0.1. In the case of EC30 and Europarl-8, the batch size is set as 8,196 tokens, accumulating gradients 20 and 8 times, respectively. Also, data from different language pairs are sampled with a temperature of 5.0 and 2.0, respectively. The same temperature is applied to both BPE building and MMT training periods. We train all models with an early-stopping strategy\footnote{Patience is set to \{10, 20\}, i.e., training stops if performance on the validation set does not improve for the last \{10, 20\} checkpoints, with 1,000 steps between checkpoints.} and evaluate by using the best checkpoint as selected based on the loss on the development set. 

For fine-tuning, all parameters are kept the same as those in training, except for 1) we set batch accumulation as 1 in Section~\ref{analysis} as extremely small fine-tuning data is used, and 2) we set patience as 3 for quick experiments.

Note that we use 4 A6000 GPU cards for English-centric training with FP16 optimization, which means the actual batch size is also 4 times bigger. For fine-tuning, we use a single A6000 GPU card.

\begin{table*}[t]
\centering
\scalebox{0.6}{
\begin{tabular}{l|ccc|ccc|ccc|ccc|ccc}
\hline\hline
\multirow{2}{*}{\textbf{Resource}} & \multicolumn{3}{c|}{\textbf{Germanic}} & \multicolumn{3}{c|}{\textbf{Romance}} & \multicolumn{3}{c|}{\textbf{Slavic}} & \multicolumn{3}{c|}{\textbf{Indo-Aryan}} & \multicolumn{3}{c}{\textbf{Afo-Asiatic}} \\
\cline{2-16} & ISO & Language & Script & ISO & Language & Script & ISO & Language & Script & ISO & Language & Script & ISO & Language & Script \\
\hline
\multirow{2}{*}{High (5M)} & DE & German & Latin & FR & French & Latin & RU & Russian & Cyrillic & HI & Hindi & Devanagari & AR & Arabic & Arabic \\
& NL & Dutch & Latin & ES & Spain & Latin & CS & Czech & Latin & BN & Bengali & Bengali & HE & Hebrew & Hebrew \\
\hline
\multirow{2}{*}{Med (1M)}
& SV & Swedish & Latin & IT & Italian & Latin & PL & Polish & Latin & KN & Kannada & Devanagari & MT & Maltese & Latin \\
& DA & Danish & Latin & PT & Portuguese & Latin & BG & Bulgarian & Cyrillic & MR & Marathi & Devanagari & HA & Hausa* & Latin \\
\hline
\multirow{2}{*}{Low (100K)}
& AF & Afrikaans & Latin & RO & Romanian & Latin & UK & Ukrainian & Cyrillic & SD & Sindhi & Arabic & TI & Tigrinya & Ethiopic \\
& LB & Luxembourgish & Latin & OC & Occitan & Latin & SR & Serbian & Latin & GU & Gujarati & Devanagari & AM & Amharic & Ethiopic \\
\hline
\multirow{2}{*}{eLow (50K)}
& NO & Norwegian & Latin & AST & Asturian & Latin & BE & Belarusian & Cyrillic & NE & Nepali & Devanagari & KAB & Kabyle* & Latin \\
& IC & Icelandic & Latin & CA & Catalan & Latin & BS & Bosnian & Latin & UR & Urdu & Arabic & So & Somali & Latin \\
\hline
\hline
\end{tabular}}
\caption{\label{table-appendix-1} Details of the EC40 dataset. Numbers in the table represent the number of sentences, e.g., 5M denotes exactly 5,000,000 sentences. Two exceptions are Hausa and Kabyle, where the size is 334K and 18K, respectively.}
\end{table*}

\subsection{Detailed Results on EC30}\label{results_on_EC30}
We report our detailed results in 970 directions (including English-centric and zero-shot directions) on EC30 datasets for both one-tag and two-tag models. The results are measured by 3 widely used metrics, i.e., ChrF, SacreBLEU, and COMET.

Table~\ref{table-appendix-EC30-weak-ChrF}, Table~\ref{table-appendix-EC30-weak-SacreBLEU} and Table~\ref{table-appendix-EC30-weak-COMET} show the specific performance of the two-tag model in each direction measured by ChrF, SacreBLEU, and COMET, respectively. In each table, we report the corresponding performance of the baseline, boost-100, and boost-all models. We also report the corresponding results in one-tag setting in Table~\ref{table-appendix-EC30-strong-ChrF}, Table~\ref{table-appendix-EC30-strong-SacreBLEU}, and Table~\ref{table-appendix-EC30-strong-COMET}, respectively. The results grouped by resource level can be found in Table~\ref{table-EC30-zero-shot-ChrF},~\ref{table-appendix-EC30-SacreBLEU}, and ~\ref{table-appendix-EC30-COMET} for ChrF, SacreBLEU, and COMET, respectively.

We also report the influence fine-tuning brings on English-centric directions, which can be found in Table~\ref{table-EC30-EN-centric-ChrF},~\ref{table-EC30-EN-centric-SacreBLEU}, and ~\ref{table-EC30-EN-centric-COMET} for ChrF, SacreBLEU, and COMET, respectively.

\begin{figure*}[t]
  \begin{minipage}{0.33\textwidth}
    \centering
    \subfigure[Within Each Family]{\includegraphics[width=5.5cm]{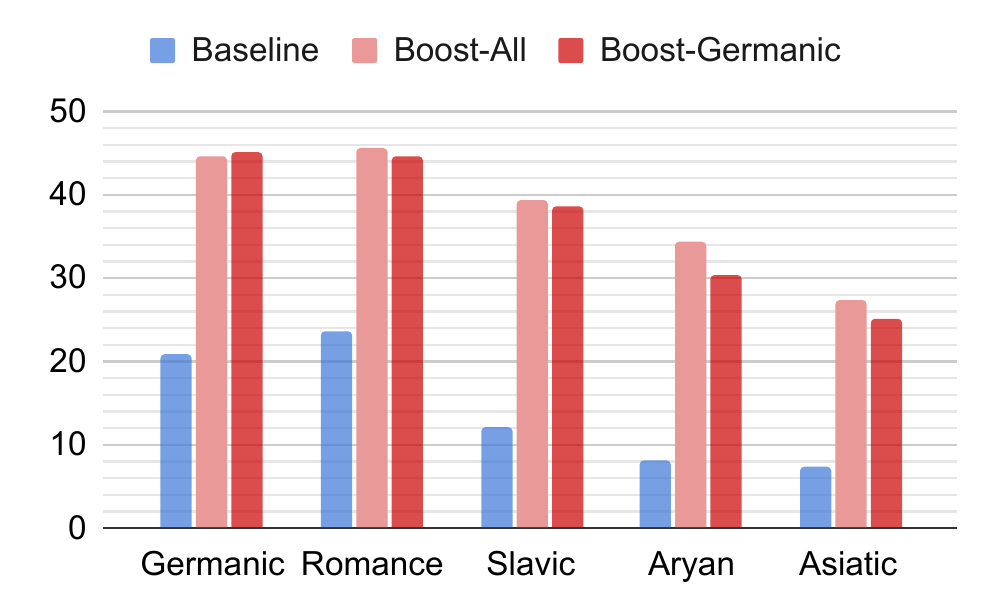}}
  \end{minipage}%
  \begin{minipage}{0.33\textwidth}
    \centering
    \subfigure[Out of Germanic]{\includegraphics[width=4.5cm]{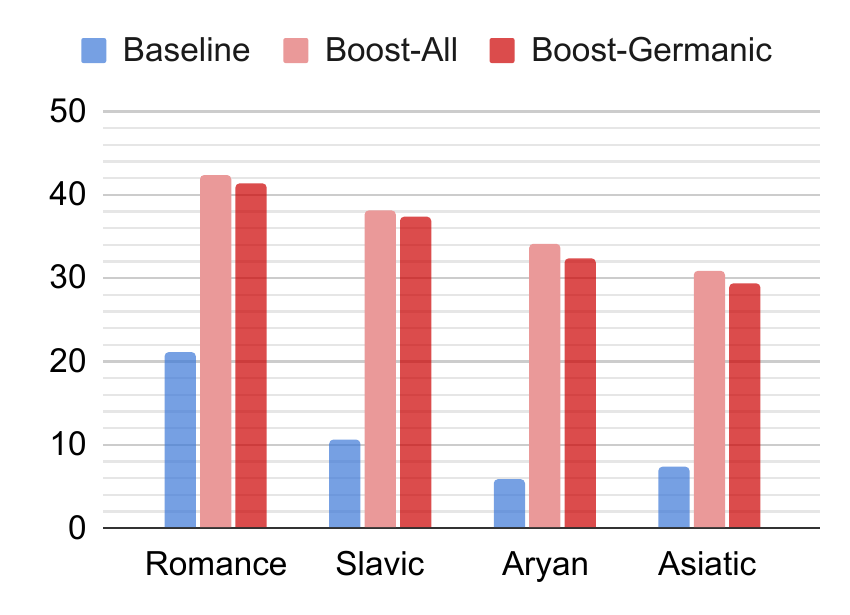}}
  \end{minipage}
    \begin{minipage}{0.33\textwidth}
    \centering
    \subfigure[Into Germanic]{\includegraphics[width=4.5cm]{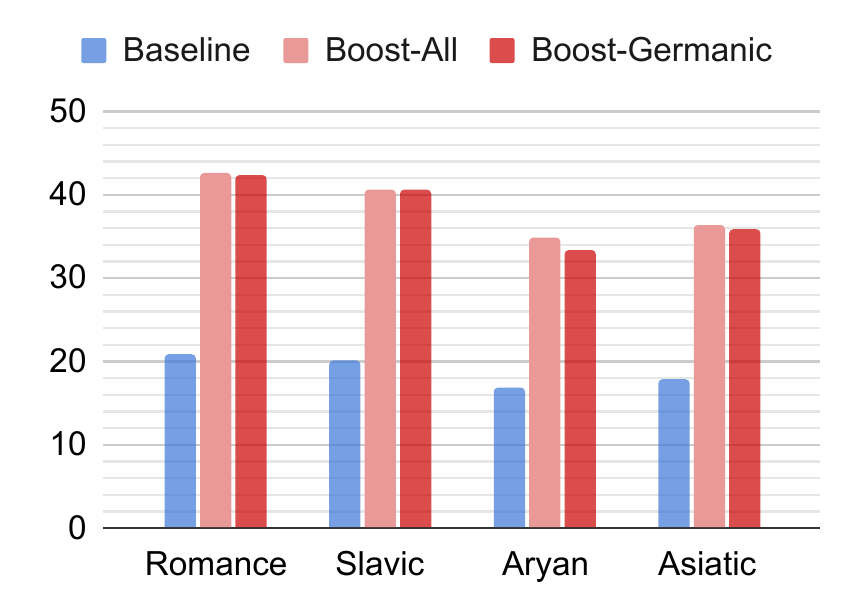}}
  \end{minipage}
  \caption{Zero-shot performance (ChrF) on EC30. Boost-All means fully fine-tuning, while Boost-Germanic means partially fine-tuning using Germanic languages. (a) shows the average performance evaluated within a specific language group, where both the source and target languages belong. (b) and (c) show the average performance in out-of-Germanic and into-Germanic directions, respectively. Detailed results are provided in Table~\ref{table-more-data-more-direction}.}
  \label{fig:fig6}
\end{figure*}

\subsection{More Data or More Directions?}
\subsubsection{Limited Fine-tuning Direction Set}\label{Limied_Direction_Set}
To further investigate the surprising boosting effects that partial directional data brings, we limit the fine-tuning direction set to fall in a specific family and check the corresponding influence across language families. Here, we limit the fine-tuning set within \textit{Germanic} (including English) and also use NTREX to build pairwise samples to cover all of the possible 42 translation directions.

Figure~\ref{fig:fig6} summarizes the zero-shot performance across language groups. It is easy to see that even when limiting fine-tuning to a specific language family (Boost-Germanic), the overall performance remains comparable to full fine-tuning (Boost-All). More specifically, Boost-Germanic achieves a slight improvement over Boost-All in Germanic directions, meanwhile slightly lagging behind in all other groups, which is also intuitive. However, the gap between the two settings is still small. This finding further demonstrates the insensitivity of the directional data during fine-tuning. Detailed results, including those in other metrics, are provided in Table~\ref{table-more-data-more-direction}.

\subsubsection{Detailed Results: More Data or More Directions?}
Table~\ref{table-more-data-more-direction} shows the detailed results when fine-tuning with Germanic data and all of the NTREX data.

\subsection{Detailed Results: How Close to the Upper Bound?}~\label{appendix_how_close}
In Table~\ref{table-appendix-Europarl-weak-ChrF},~\ref{table-appendix-Europarl-weak-Comet} and ~\ref{table-appendix-Europarl-weak-SacreBLEU}, we show the detailed results for the baseline, boosted, and upper bound models on the Europarl-8 dataset in the metric of ChrF, COMET, and SacreBLEU, respectively.

\subsection{Number Pairs}
\begin{figure*}
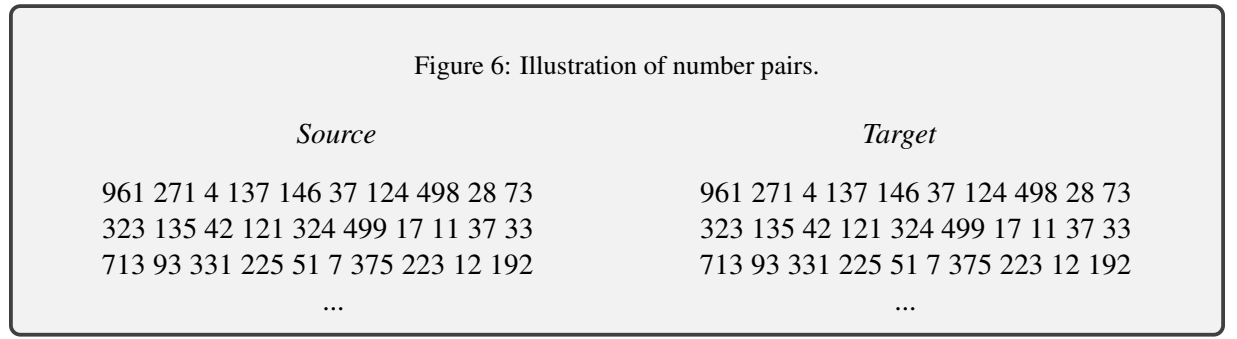

\begin{tcolorbox}
\caption{Illustration of number pairs.}
\label{fig:number_pairs}
\begin{center} 
\textit{Source} \hspace{6.2cm} \textit{Target} \\
\vspace{0.3cm}
961 271 4 137 146 37 124 498 28 73 \hspace{2cm} 961 271 4 137 146 37 124 498 28 73 \\
323 135 42 121 324 499 17 11 37 33 \hspace{2cm} 323 135 42 121 324 499 17 11 37 33 \\
713 93 331 225 51 7 375 223 12 192 \hspace{2cm} 713 93 331 225 51 7 375 223 12 192 \\
... \hspace{7cm} ... \\
\end{center}
\end{tcolorbox}
\end{figure*}

The synthetic number pairs and word pairs are illustrated in Figure~\ref{fig:number_pairs}.

% Appendix Table-EC30-Weak-ChrF
\begin{table*}[t]
\centering
\scalebox{0.19}{
\begin{tabular}{c|cccccccccccccccccccccccccccccccccccc}
\hline
\hline
ChrF & en & de & nl & fr & es & ru & cs & hi & bn & ar & he & sv & da & it & pt & pl & bg & kn & mr & mt & ha & af & am & gu & lb & oc & ro & sd & sr & ti & uk \\
\hline
en &  -  & 57.3/56.6/57.0 & 52.2/51.8/51.6 & 63.0/61.4/61.9 & 49.7/49.1/49.3 & 48.6/47.1/47.3 & 50.8/49.2/49.7 & 54.3/53.4/54.2 & 50.6/50.6/50.5 & 46.1/46.1/46.1 & 51.9/51.3/51.9 & 60.3/58.2/58.7 & 60.3/58.7/59.4 & 52.1/51.5/51.5 & 59.6/57.9/58.6 & 43.6/43.4/43.6 & 61.3/59.6/60.3 & 51.5/50.6/51.0 & 44.8/46.5/45.8 & 65.8/64.7/65.6 & 40.1/43.7/42.7 & 62.8/62.0/62.7 & 29.3/31.2/30.9 & 46.5/47.6/47.0 & 42.6/45.8/44.7 & 52.0/52.9/53.2 & 51.6/52.5/52.1 & 18.8/23.3/22.4 & 51.3/49.7/49.6 & 22.4/23.7/22.8 & 47.3/45.9/46.5 \\
de & 62.6/60.8/61.2 &  -  & 20.8/44.6/44.2 & 22.0/48.0/46.7 & 19.9/41.2/40.8 & 3.0/38.5/37.9 & 16.2/39.5/38.9 & 3.0/38.2/37.7 & 3.3/38.4/37.8 & 3.6/33.7/32.6 & 3.1/35.5/34.8 & 21.1/46.5/45.7 & 21.7/46.8/45.9 & 21.2/41.7/40.8 & 20.7/43.7/43.0 & 16.6/35.8/35.2 & 3.7/45.5/44.8 & 7.6/38.3/39.1 & 8.2/36.3/35.2 & 17.2/42.7/41.2 & 14.5/31.8/29.9 & 20.8/45.8/44.6 & 1.0/19.9/18.1 & 4.4/34.5/33.2 & 19.3/41.3/39.9 & 21.0/39.3/38.8 & 20.3/41.2/39.2 & 1.5/18.5/16.2 & 16.5/37.7/36.5 & 1.7/17.8/15.5 & 2.5/35.5/34.3 \\
nl & 54.9/54.0/54.2 & 19.0/43.4/42.9 &  -  & 21.6/44.0/43.2 & 19.5/39.4/38.3 & 2.8/35.1/34.8 & 16.0/35.3/34.9 & 2.9/36.0/35.8 & 3.4/36.8/36.3 & 3.6/31.5/30.7 & 3.2/32.8/32.2 & 20.5/41.8/41.0 & 21.4/42.3/41.6 & 20.7/39.2/38.3 & 20.2/40.7/40.2 & 16.3/33.6/33.3 & 3.4/41.4/40.8 & 8.4/36.5/37.3 & 8.4/34.9/33.6 & 16.8/39.7/38.7 & 14.4/31.4/29.6 & 20.6/43.9/43.2 & 1.0/18.9/17.3 & 4.7/32.5/31.2 & 18.2/35.2/33.9 & 20.7/37.3/36.6 & 19.8/37.8/36.1 & 1.5/17.8/16.1 & 16.2/34.7/33.4 & 1.9/16.8/14.8 & 2.5/32.6/31.6 \\
\hline
\hline
\end{tabular}}
\caption{\label{table-appendix-EC30-weak-ChrF} The ChrF performance on EC30 for the baseline, boost-100, and boost-all models in the two-tag fashion, respectively. 970 directions of results are shown, including both English-centric and zero-shot ones. Overall, in all 870 zero-shot directions, the boost-100 model achieves better performance compared to the baseline. Moreover, in 845 out of 870 zero-shot directions, the gain exceeds 10.0.} 
\end{table*}

% Appendix Table-EC30-Weak-SacreBLEU
\begin{table*}[t]
\centering
\scalebox{0.19}{
\begin{tabular}{c|cccccccccccccccccccccccccccccccccccc}
\hline
\hline
SacreBLEU & en & de & nl & fr & es & ru & cs & hi & bn & ar & he & sv & da & it & pt & pl & bg & kn & mr & mt & ha & af & am & gu & lb & oc & ro & sd & sr & ti & uk \\
\hline
en &  -  & 31.0/30.1/30.1 & 23.4/22.7/22.9 & 40.6/39.5/39.0 & 23.1/22.9/23.2 & 22.4/21.1/21.6 & 25.0/23.5/23.4 & 38.6/38.5/37.5 & 32.0/32.1/31.7 & 19.2/18.7/19.4 & 24.6/23.7/22.2 & 35.8/33.5/33.6 & 35.9/34.7/34.8 & 24.4/23.7/23.7 & 34.4/33.2/33.1 & 16.2/15.9/16.3 & 35.7/34.1/33.9 & 33.6/34.1/30.0 & 26.6/27.4/27.5 & 37.9/37.7/37.6 & 14.7/16.1/16.8 & 36.2/34.7/36.2 & 7.5/9.0/9.4 & 26.8/26.8/26.7 & 13.9/14.5/15.9 & 24.9/24.8/24.9 & 24.8/26.9/27.6 & 4.2/4.2/4.5 & 23.7/22.7/22.7 & 4.2/4.9/5.5 & 21.2/19.8/19.7 \\
de & 37.6/36.3/35.3 &  -  & 2.5/16.4/16.5 & 2.9/21.7/23.4 & 2.2/15.5/16.5 & 1.2/13.1/13.8 & 1.9/14.9/15.5 & 1.4/20.9/21.4 & 2.0/18.9/19.9 & 0.8/9.4/10.2 & 0.8/10.5/10.5 & 2.4/20.1/20.8 & 2.9/21.7/22.5 & 2.5/15.8/16.1 & 2.4/17.6/18.6 & 2.1/11.0/11.9 & 1.7/20.3/20.1 & 4.3/21.1/21.0 & 3.8/16.1/17.5 & 2.0/15.1/16.3 & 1.9/8.4/9.9 & 2.6/19.4/20.8 & 0.4/3.8/4.5 & 2.1/14.3/15.6 & 2.2/12.1/13.2 & 2.4/13.3/13.8 & 2.3/16.0/17.6 & 0.5/2.0/2.6 & 1.4/12.4/13.8 & 0.2/2.4/3.5 & 0.8/11.3/12.0 \\
nl & 28.2/27.2/27.5 & 2.6/15.5/16.0 &  -  & 3.1/17.8/18.9 & 2.0/13.8/14.5 & 0.9/10.0/10.5 & 1.9/11.6/12.1 & 1.3/19.1/19.5 & 2.0/16.8/18.2 & 0.6/7.2/8.2 & 0.9/8.5/8.6 & 2.3/15.2/15.8 & 2.8/16.0/17.2 & 2.3/12.9/13.4 & 2.4/14.7/15.4 & 2.1/9.6/9.6 & 1.4/16.0/16.4 & 5.0/19.4/19.8 & 3.4/15.2/16.3 & 1.8/12.3/13.7 & 1.7/7.9/8.7 & 2.5/15.6/17.3 & 0.4/2.9/3.7 & 2.3/13.5/14.5 & 1.9/7.9/9.3 & 2.5/11.4/11.7 & 2.3/12.8/14.4 & 0.5/1.7/2.4 & 1.2/9.5/10.4 & 0.2/2.0/2.9 & 0.7/8.6/9.6 \\
\hline
\hline
\end{tabular}}
\caption{\label{table-appendix-EC30-weak-SacreBLEU} The SacreBLEU performance on EC30 for the baseline, boost-100, and boost-all models in the two-tag fashion, respectively. 970 directions of results are shown, including English-centric and zero-shot ones. Overall, in all 870 zero-shot directions, the boost-100 model achieves better performance compared to the baseline. Moreover, in 702 out of 870 zero-shot directions, the gain exceeds 5.0.} 
\end{table*}

% Appendix Table-EC30-Weak-COMET
\begin{table*}[]
\centering
\scalebox{0.19}{
\begin{tabular}{c|cccccccccccccccccccccccccccccccccccc}
\hline
\hline
COMET & en & de & nl & fr & es & ru & cs & hi & bn & ar & he & sv & da & it & pt & pl & bg & kn & mr & mt & ha & af & am & gu & lb & oc & ro & sd & sr & ti & uk \\
\hline
en &  -  & 82.8/82.1/82.2 & 84.2/83.7/84.0 & 82.8/81.5/82.0 & 82.0/81.6/81.5 & 81.9/81.4/81.3 & 84.5/83.9/84.0 & 78.1/78.4/78.3 & 84.5/84.9/84.8 & 79.0/79.0/79.1 & 83.3/83.8/83.7 & 85.4/84.2/84.2 & 85.3/84.4/84.4 & 84.0/83.2/83.4 & 85.1/84.5/84.4 & 82.0/81.5/81.6 & 87.2/86.9/86.6 & 87.5/87.5/86.0 & 74.4/75.2/75.5 & 70.8/71.0/70.8 & 68.7/72.1/73.4 & 84.0/83.3/83.0 & 78.2/80.9/81.0 & 85.1/85.4/85.3 & 53.0/53.7/54.0 & 66.0/65.0/66.0 & 80.2/78.5/79.3 & 47.5/48.9/51.8 & 82.1/80.5/80.7 & 80.5/81.1/81.9 & 80.7/79.7/80.0 \\
de & 86.4/85.9/85.6 &  -  & 61.0/78.6/78.1 & 62.5/70.9/71.1 & 64.7/71.3/71.8 & 45.8/72.5/73.1 & 56.1/75.6/77.4 & 36.4/68.1/68.2 & 36.4/77.6/78.1 & 46.5/68.1/69.7 & 53.7/71.4/72.5 & 66.8/76.5/76.8 & 66.0/76.5/76.8 & 61.8/73.3/74.1 & 65.7/73.5/74.5 & 55.2/73.2/74.4 & 49.5/76.6/76.8 & 42.2/82.1/81.6 & 37.5/66.6/67.2 & 36.1/63.1/63.8 & 47.8/65.0/66.9 & 62.7/75.3/75.3 & 41.7/66.8/69.1 & 41.5/75.9/76.9 & 41.4/53.0/53.5 & 55.3/58.1/58.6 & 59.4/68.1/70.1 & 39.4/39.6/45.2 & 55.5/69.5/71.3 & 32.3/73.4/75.6 & 46.3/67.5/69.4 \\
nl & 84.7/84.5/84.1 & 55.2/74.7/74.8 &  -  & 61.0/70.6/70.7 & 63.5/71.3/71.3 & 44.6/71.6/72.4 & 54.6/74.5/75.1 & 35.6/68.0/68.2 & 36.2/77.8/78.3 & 45.7/66.9/68.7 & 52.4/71.1/72.2 & 65.0/74.7/74.8 & 64.6/75.2/75.3 & 60.7/72.2/73.1 & 64.2/72.6/73.1 & 54.8/72.6/73.2 & 48.2/75.8/76.1 & 43.3/82.4/81.9 & 37.9/66.4/67.1 & 35.5/62.1/62.7 & 46.9/64.6/66.8 & 60.9/78.0/77.2 & 40.6/66.7/69.0 & 41.4/76.3/77.1 & 40.1/50.6/51.2 & 54.2/57.0/57.3 & 57.9/67.0/69.0 & 38.4/39.8/44.5 & 54.1/68.2/69.3 & 32.3/72.9/75.5 & 45.3/66.9/68.4 \\
\hline
\hline
\end{tabular}}
\caption{\label{table-appendix-EC30-weak-COMET} The COMET performance on EC30 for the baseline, boost-100, and boost-all models in the two-tag fashion, respectively. 970 directions of results are shown, including English-centric and zero-shot ones. Overall, in 869 out of 870 zero-shot directions, the boost-100 model achieves better performance compared to the baseline. Moreover, in 636 out of 870 zero-shot directions, the gain exceeds 10.0.} 
\end{table*}

% Appendix Table-EC30-strong-ChrF
\begin{table*}[t]
\centering
\scalebox{0.19}{
\begin{tabular}{c|cccccccccccccccccccccccccccccccccccc}
\hline
\hline
ChrF & en & de & nl & fr & es & ru & cs & hi & bn & ar & he & sv & da & it & pt & pl & bg & kn & mr & mt & ha & af & am & gu & lb & oc & ro & sd & sr & ti & uk \\
\hline
en &  -  & 57.9/57.0/56.6 & 52.1/51.6/51.8 & 62.9/61.9/61.4 & 49.6/49.3/49.1 & 48.5/47.3/47.1 & 51.2/49.7/49.2 & 54.4/54.2/53.4 & 50.8/50.5/50.6 & 46.9/46.1/46.1 & 52.2/51.9/51.3 & 60.3/58.7/58.2 & 60.5/59.4/58.7 & 51.9/51.5/51.5 & 59.5/58.6/57.9 & 43.9/43.6/43.4 & 61.3/60.3/59.6 & 52.0/51.0/50.6 & 44.9/45.8/46.5 & 65.7/65.6/64.7 & 39.5/42.7/43.7 & 63.9/62.7/62.0 & 28.9/30.9/31.2 & 46.9/47.0/47.6 & 43.1/44.7/45.8 & 53.8/53.2/52.9 & 51.9/52.1/52.5 & 19.5/22.4/23.3 & 51.5/49.6/49.7 & 22.3/22.8/23.7 & 47.5/46.5/45.9 \\
de & 61.9/61.2/60.8 &  -  & 41.5/44.2/44.6 & 41.4/46.7/48.0 & 35.0/40.8/41.2 & 26.7/37.9/38.5 & 25.3/38.9/39.5 & 31.8/37.7/38.2 & 31.2/37.8/38.4 & 26.2/32.6/33.7 & 19.9/34.8/35.5 & 42.3/45.7/46.5 & 41.3/45.9/46.8 & 38.5/40.8/41.7 & 40.0/43.0/43.7 & 30.8/35.2/35.8 & 42.2/44.8/45.5 & 36.8/39.1/38.3 & 28.9/35.2/36.3 & 28.9/41.2/42.7 & 16.9/29.9/31.8 & 35.1/44.6/45.8 & 2.0/18.1/19.9 & 29.5/33.2/34.5 & 26.8/39.9/41.3 & 30.4/38.8/39.3 & 32.4/39.2/41.2 & 3.5/16.2/18.5 & 32.6/36.5/37.7 & 11.7/15.5/17.8 & 29.9/34.3/35.5 \\
nl & 54.4/54.2/54.0 & 40.9/42.9/43.4 &  -  & 40.1/43.2/44.0 & 36.1/38.3/39.4 & 28.3/34.8/35.1 & 25.8/34.9/35.3 & 31.5/35.8/36.0 & 31.7/36.3/36.8 & 27.2/30.7/31.5 & 21.3/32.2/32.8 & 39.3/41.0/41.8 & 38.4/41.6/42.3 & 36.8/38.3/39.2 & 38.4/40.2/40.7 & 30.1/33.3/33.6 & 39.3/40.8/41.4 & 35.4/37.3/36.5 & 28.1/33.6/34.9 & 29.9/38.7/39.7 & 17.7/29.6/31.4 & 39.4/43.2/43.9 & 3.1/17.3/18.9 & 28.3/31.2/32.5 & 25.1/33.9/35.2 & 32.2/36.6/37.3 & 32.8/36.1/37.8 & 5.6/16.1/17.8 & 31.4/33.4/34.7 & 12.1/14.8/16.8 & 29.6/31.6/32.6 \\
\hline
\hline
\end{tabular}}
\caption{\label{table-appendix-EC30-strong-ChrF} The ChrF performance on EC30 for the baseline, boost-100, and boost-all models in the one-tag fashion, respectively. 970 directions of results are shown, including English-centric and zero-shot ones. Overall, in all 870 zero-shot directions, the boost-100 model achieves better performance compared to the baseline. Moreover, in 564 out of 870 zero-shot directions, the gain exceeds 5.0.} 
\end{table*}

% Appendix Table-EC30-strong-SacreBLEU
\begin{table*}[t]
\centering
\scalebox{0.19}{
\begin{tabular}{c|cccccccccccccccccccccccccccccccccccc}
\hline
\hline
SacreBLEU & en & de & nl & fr & es & ru & cs & hi & bn & ar & he & sv & da & it & pt & pl & bg & kn & mr & mt & ha & af & am & gu & lb & oc & ro & sd & sr & ti & uk \\
\hline
en &  -  & 31.0/30.1/30.1 & 23.4/22.7/22.9 & 40.6/39.5/39.0 & 23.1/22.9/23.2 & 22.4/21.1/21.6 & 25.0/23.5/23.4 & 38.6/38.5/37.5 & 32.0/32.1/31.7 & 19.2/18.7/19.4 & 24.6/23.7/22.2 & 35.8/33.5/33.6 & 35.9/34.7/34.8 & 24.4/23.7/23.7 & 34.4/33.2/33.1 & 16.2/15.9/16.3 & 35.7/34.1/33.9 & 33.6/34.1/30.0 & 26.6/27.4/27.5 & 37.9/37.7/37.6 & 14.7/16.1/16.8 & 36.2/34.7/36.2 & 7.5/9.0/9.4 & 26.8/26.8/26.7 & 13.9/14.5/15.9 & 24.9/24.8/24.9 & 24.8/26.9/27.6 & 4.2/4.2/4.5 & 23.7/22.7/22.7 & 4.2/4.9/5.5 & 21.2/19.8/19.7 \\
de & 37.6/36.3/35.3 &  -  & 2.5/16.4/16.5 & 2.9/21.7/23.4 & 2.2/15.5/16.5 & 1.2/13.1/13.8 & 1.9/14.9/15.5 & 1.4/20.9/21.4 & 2.0/18.9/19.9 & 0.8/9.4/10.2 & 0.8/10.5/10.5 & 2.4/20.1/20.8 & 2.9/21.7/22.5 & 2.5/15.8/16.1 & 2.4/17.6/18.6 & 2.1/11.0/11.9 & 1.7/20.3/20.1 & 4.3/21.1/21.0 & 3.8/16.1/17.5 & 2.0/15.1/16.3 & 1.9/8.4/9.9 & 2.6/19.4/20.8 & 0.4/3.8/4.5 & 2.1/14.3/15.6 & 2.2/12.1/13.2 & 2.4/13.3/13.8 & 2.3/16.0/17.6 & 0.5/2.0/2.6 & 1.4/12.4/13.8 & 0.2/2.4/3.5 & 0.8/11.3/12.0 \\
nl & 28.2/27.2/27.5 & 2.6/15.5/16.0 &  -  & 3.1/17.8/18.9 & 2.0/13.8/14.5 & 0.9/10.0/10.5 & 1.9/11.6/12.1 & 1.3/19.1/19.5 & 2.0/16.8/18.2 & 0.6/7.2/8.2 & 0.9/8.5/8.6 & 2.3/15.2/15.8 & 2.8/16.0/17.2 & 2.3/12.9/13.4 & 2.4/14.7/15.4 & 2.1/9.6/9.6 & 1.4/16.0/16.4 & 5.0/19.4/19.8 & 3.4/15.2/16.3 & 1.8/12.3/13.7 & 1.7/7.9/8.7 & 2.5/15.6/17.3 & 0.4/2.9/3.7 & 2.3/13.5/14.5 & 1.9/7.9/9.3 & 2.5/11.4/11.7 & 2.3/12.8/14.4 & 0.5/1.7/2.4 & 1.2/9.5/10.4 & 0.2/2.0/2.9 & 0.7/8.6/9.6 \\
\hline
\hline
\end{tabular}}
\caption{\label{table-appendix-EC30-strong-SacreBLEU} The SacreBLEU performance on EC30 for the baseline, boost-100, and boost-all models in the one-tag fashion, respectively. 970 directions of results are shown, including English-centric and zero-shot ones. Overall, in 869 out of 870 zero-shot directions, the boost-100 model achieves better performance compared to the baseline. Moreover, in 423 out of 870 zero-shot directions, the gain exceeds 3.0.} 
\end{table*}

% Appendix Table-EC30-strong-COMET
\begin{table*}[]
\centering
\scalebox{0.19}{
\begin{tabular}{c|cccccccccccccccccccccccccccccccccccc}
\hline
\hline
COMET & en & de & nl & fr & es & ru & cs & hi & bn & ar & he & sv & da & it & pt & pl & bg & kn & mr & mt & ha & af & am & gu & lb & oc & ro & sd & sr & ti & uk \\
\hline
en &  -  & 82.5/82.1/81.4 & 84.0/83.8/83.6 & 82.5/81.6/81.3 & 81.6/80.9/80.8 & 81.8/81.1/80.9 & 84.4/83.6/83.1 & 77.6/78.0/77.8 & 84.3/84.5/84.5 & 78.9/78.7/78.8 & 82.9/83.3/83.4 & 84.9/83.9/83.6 & 85.2/84.2/83.6 & 83.5/82.7/82.5 & 84.5/84.1/83.5 & 81.8/81.4/81.1 & 86.9/86.5/86.2 & 87.4/87.4/86.5 & 74.4/75.1/75.3 & 70.6/70.9/70.5 & 66.7/71.6/73.4 & 83.4/82.8/82.3 & 75.2/79.3/80.5 & 84.8/85.2/85.4 & 52.9/53.9/54.1 & 66.1/66.0/66.6 & 79.7/77.8/77.5 & 47.3/51.9/52.3 & 81.3/79.7/79.8 & 80.6/81.2/81.7 & 80.3/79.6/79.1 \\
de & 85.8/85.5/85.2 &  -  & 72.5/76.5/76.5 & 63.7/68.3/69.1 & 64.0/69.4/69.8 & 58.6/70.1/70.7 & 57.8/72.6/73.9 & 61.4/66.6/66.7 & 70.2/77.2/77.0 & 58.1/64.9/66.3 & 57.2/67.4/68.6 & 70.1/73.5/74.3 & 69.6/73.4/74.2 & 66.6/69.6/71.2 & 67.8/70.8/71.7 & 63.6/69.6/70.8 & 69.1/73.0/74.2 & 79.4/81.3/79.8 & 59.6/65.3/66.0 & 48.5/62.4/62.7 & 46.5/60.5/63.6 & 66.3/71.8/72.1 & 41.1/63.5/66.8 & 70.1/74.9/75.9 & 45.4/52.9/53.4 & 55.6/57.3/57.8 & 59.0/63.5/65.6 & 38.0/41.5/44.5 & 60.4/65.1/66.7 & 63.9/71.5/75.2 & 59.2/63.8/65.6 \\
nl & 84.2/84.0/83.8 & 69.2/71.7/72.4 &  -  & 63.7/67.1/67.6 & 65.1/68.2/69.0 & 60.4/69.0/69.9 & 57.6/70.3/71.7 & 62.7/66.5/66.6 & 72.6/77.5/77.5 & 59.7/64.3/65.6 & 57.8/66.8/68.7 & 68.8/71.4/72.1 & 68.9/72.2/72.5 & 66.7/69.1/70.0 & 67.3/69.8/70.5 & 63.3/69.4/69.6 & 69.2/72.6/73.7 & 79.9/81.5/80.5 & 59.9/65.7/66.1 & 51.4/61.7/61.9 & 46.2/61.4/63.9 & 73.5/76.1/75.6 & 41.0/63.5/66.7 & 70.9/74.6/75.9 & 44.7/50.4/50.9 & 55.4/56.8/57.4 & 59.8/62.9/64.5 & 37.0/42.0/44.5 & 58.9/62.8/65.2 & 66.0/71.5/74.6 & 60.5/63.9/65.4 \\
\hline
\hline
\end{tabular}}
\caption{\label{table-appendix-EC30-strong-COMET} The COMET performance on EC30 for the baseline, boost-100, and boost-all models in the one-tag fashion, respectively. 970 directions of results are shown, including English-centric and zero-shot ones. Overall, in 866 out of 870 zero-shot directions, the boost-100 model achieves better performance compared to the baseline. Moreover, in 501 out of 870 zero-shot directions, the gain exceeds 5.0.} 
\end{table*}

% EC30-SacreBLEU (Zero-shot)
\begin{table*}[t]
\centering
\scalebox{0.95}{
\begin{tabular}{l|l|ccccccccc|c}
\hline\hline
& Model &H-H & H-M & H-L & M-H & M-M & M-L & L-H & L-M & L-L & AVG \\ \hline
\multirow{5}{*}{\rotatebox{90}{Two-Tag}} & Baseline &1.5 & 2.2 & 1.3 & 1.8 & 2.4 & 1.3 & 2.6 & 3.1 & 1.3 & 2.0 \\
& Boost-100 &13.4 & 14.2 & 9.0 & 13.8 & 14.2 & 9.2 & 10.8 & 11.2 & 6.8 & 11.4 \\
& Boost-All &14.0 & 14.9 & 10.0 & 14.4 & 14.9 & 10.1 & 11.6 & 12.1 & 7.8 & 12.2 \\ \cdashline{2-12}
& $\Delta$-100  &+11.9 & +12.0 & +7.7 & +12.0 & +11.8 & +7.9 & +8.2 & +8.1 & +5.5 & +9.4 \\
& $\Delta$-All  &+12.5 & +12.7 & +8.7 & +12.6 & +12.5 & +8.8 & +9.0 & +9.0 & +6.5 & +10.2 \\
\hline
\multirow{5}{*}{\rotatebox{90}{One-Tag}} & Baseline &8.4 & 9.0 & 4.6 & 8.8 & 8.9 & 4.7 & 6.4 & 6.7 & 3.1 & 6.7 \\
& Boost-100 &12.4 & 12.8 & 8.0 & 12.7 & 12.6 & 8.1 & 9.5 & 9.7 & 5.7 & 10.2 \\
& Boost-All &12.9 & 13.4 & 9.0 & 13.2 & 13.3 & 9.1 & 10.4 & 10.6 & 6.8 & 10.9 \\ \cdashline{2-12}
& $\Delta$-100  &+4.0 & +3.8 & +3.4 & +3.9 & +3.7 & +3.4 & +3.1 & +3.0 & +2.6 & +3.5 \\
& $\Delta$-All  &+4.5 & +4.4 & +4.4 & +4.4 & +4.4 & +4.4 & +4.0 & +3.9 & +3.7 & +4.2 \\
\hline
\hline
\end{tabular}}
\caption{\label{table-appendix-EC30-SacreBLEU} Zero-shot performance (SacreBLEU) on the EC30 dataset (61M sentence pairs) with two language tag strategies, grouped by \textbf{H}igh-, \textbf{M}edium, and \textbf{L}ow-resource, respectively. $\Delta$-100 and $\Delta$-All mean the corresponding performance changes compared with the baselines.}
\end{table*}

% EC30-SacreBLEU (English-centric)
\begin{table*}[t]
\centering
\scalebox{0.95}{
\begin{tabular}{l|l|cc|cc|cc|ccc}
\hline\hline
\multirow{2}{*}{} & \multirow{2}{*}{Model} & \multicolumn{2}{c|}{High} & \multicolumn{2}{c|}{Medium} & \multicolumn{2}{c|}{Low} & \multicolumn{3}{c}{Overall} \\ 
\cline{3-11} 
& & EN-X & X-EN & EN-X & X-EN & EN-X & X-EN & EN-X & X-EN & AVG \\ \hline
\multirow{5}{*}{\rotatebox{90}{Two-Tag}} & Baseline & 28.0 & 31.4 & 29.5 & 31.4 & 18.7 & 25.8 & 25.4 & 29.5 & 27.5 \\
& Boost-100 & 27.3 & 29.9 & 29.0 & 31.1 & 18.8 & 25.0 & 25.0 & 28.7 & 26.9 \\
& Boost-All & 27.1 & 29.7 & 28.7 & 31.1 & 19.3 & 25.8 & 25.0 & 28.9 & 27.0 \\ \cdashline{2-11}
& $\Delta$-100 & -0.7 & -1.5 & -0.5 & -0.3 & +0.1 & -0.8 & -0.4 & -0.8 & -0.6 \\
& $\Delta$-All & -0.9 & -1.7 & -0.8 & -0.3 & +0.6 & 0.0 & -0.4 & -0.6 & -0.5 \\ \hline
\multirow{5}{*}{\rotatebox{90}{One-Tag}} & Baseline & 28.4 & 30.9 & 29.8 & 30.9 & 19.5 & 25.3 & 25.9 & 29.0 & 27.5 \\
& Boost-100 & 27.6 & 30.5 & 29.6 & 31.4 & 19.4 & 25.5 & 25.5 & 29.1 & 27.3 \\
& Boost-All & 27.0 & 30.0 & 28.7 & 31.2 & 19.6 & 25.5 & 25.1 & 28.9 & 27.0 \\ \cdashline{2-11}
& $\Delta$-100 & -0.8 & -0.4 & -0.2 & +0.5 & -0.1 & +0.2 & -0.4 & +0.1 & -0.2 \\
& $\Delta$-All & -1.4 & -0.9 & -1.1 & +0.3 & +0.1 & +0.2 & -0.8 & -0.1 & -0.5 \\ \hline
\end{tabular}}
\caption{\label{table-EC30-EN-centric-SacreBLEU} English-centric performance (SacreBLEU) on the EC30 dataset (61M sentence pairs). EN-X and X-EN denote the average out-of- and into-English translation performance on each resource group, respectively.}
\end{table*}

% EC30-COMET (Zero-shot)
\begin{table*}[t]
\centering
\scalebox{0.95}{
\begin{tabular}{l|l|ccccccccc|c}
\hline\hline
& Model & H-H & H-M & H-L & M-H & M-M & M-L & L-H & L-M & L-L & AVG \\ \hline
\multirow{5}{*}{\rotatebox{90}{Two-Tag}} & Baseline & 51.0 & 52.4 & 47.3 & 48.6 & 49.9 & 45.0 & 46.4 & 49.3 & 43.6 & 48.2 \\
& Boost-100 & 70.6 & 71.2 & 64.0 & 69.1 & 69.7 & 63.1 & 63.1 & 64.8 & 58.7 & 66.0 \\
& Boost-All & 71.3 & 71.8 & 65.7 & 69.9 & 70.4 & 64.8 & 64.4 & 66.0 & 60.9 & 67.3 \\ \cdashline{2-12}
& $\Delta$-100  & +19.6 & +18.8 & +16.7 & +20.5 & +19.8 & +18.1 & +16.7 & +15.5 & +15.1 & +17.8 \\
& $\Delta$-All  & +20.3 & +19.4 & +18.4 & +21.3 & +20.5 & +19.8 & +18.0 & +16.7 & +17.3 & +19.1 \\
\hline
\multirow{5}{*}{\rotatebox{90}{One-Tag}} & Baseline & 61.2 & 62.6 & 54.4 & 58.8 & 60.2 & 52.2 & 51.9 & 54.9 & 47.8 & 56.0 \\
& Boost-100 & 68.1 & 68.8 & 61.7 & 66.5 & 67.1 & 60.5 & 60.2 & 62.1 & 55.8 & 63.4 \\
& Boost-All & 68.9 & 69.5 & 63.6 & 67.6 & 68.0 & 62.7 & 62.1 & 63.7 & 58.7 & 65.0 \\ \cdashline{2-12}
& $\Delta$-100  & +6.9 & +6.2 & +7.3 & +7.7 & +6.9 & +8.3 & +8.3 & +7.2 & +8.0 & +7.4 \\
& $\Delta$-All  & +7.7 & +6.9 & +9.2 & +8.8 & +7.8 & +10.5 & +10.2 & +8.8 & +10.9 & +9.0 \\
\hline
\hline
\end{tabular}}
\caption{\label{table-appendix-EC30-COMET} Zero-shot performance (COMET) on the EC30 dataset (61M sentence pairs) with two language tag strategies, grouped by \textbf{H}igh-, \textbf{M}edium, and \textbf{L}ow-resource, respectively. $\Delta$-100 and $\Delta$-All mean the corresponding performance changes compared with the baselines.}
\end{table*}

% EC30-COMET (English-centric)
\begin{table*}[t]
\centering
\scalebox{0.95}{
\begin{tabular}{l|l|cc|cc|cc|ccc}
\hline\hline
\multirow{2}{*}{} & \multirow{2}{*}{Model} & \multicolumn{2}{c|}{High} & \multicolumn{2}{c|}{Medium} & \multicolumn{2}{c|}{Low} & \multicolumn{3}{c}{Overall} \\ 
\cline{3-11} 
& & EN-X & X-EN & EN-X & X-EN & EN-X & X-EN & EN-X & X-EN & AVG \\ \hline
\multirow{5}{*}{\rotatebox{90}{Two-Tag}} & Baseline & 82.3 & 83.8 & 81.0 & 80.0 & 73.7 & 73.3 & 79.0 & 79.0 & 79.0 \\
& Boost-100 & 82.0 & 83.4 & 81.1 & 81.3 & 73.7 & 74.4 & 78.9 & 79.7 & 79.3 \\
& Boost-All & 82.1 & 83.0 & 81.0 & 81.1 & 74.3 & 75.2 & 79.1 & 79.8 & 79.5 \\ \cdashline{2-11}
& $\Delta$-100 & -0.3 & -0.4 & +0.1 & +1.3 & 0.0 & +1.1 & -0.1 & +0.7 & +0.3 \\
& $\Delta$-All & -0.2 & -0.8 & 0.0 & +1.1 & +0.6 & +1.9 & +0.1 & +0.8 & +0.5 \\ \hline
\multirow{5}{*}{\rotatebox{90}{One-Tag}} & Baseline & 82.0 & 83.3 & 80.6 & 79.6 & 73.2 & 72.6 & 78.6 & 78.5 & 78.5 \\
& Boost-100 & 81.8 & 83.0 & 80.8 & 80.9 & 73.7 & 74.2 & 78.8 & 79.4 & 79.1 \\
& Boost-All & 81.6 & 82.7 & 80.6 & 80.7 & 73.9 & 74.7 & 78.7 & 79.4 & 79.0 \\ \cdashline{2-11}
& $\Delta$-100 & -0.2 & -0.3 & +0.2 & +1.3 & +0.5 & +1.6 & +0.2 & +0.9 & +0.6 \\
& $\Delta$-All & -0.4 & -0.6 & 0.0 & +1.1 & +0.7 & +2.1 & +0.1 & +0.9 & +0.5 \\ \hline
\end{tabular}}
\caption{\label{table-EC30-EN-centric-COMET} English-centric performance (COMET) on the EC30 dataset (61M sentence pairs). EN-X and X-EN denote the average out-of- and into-English translation performance on each resource group, respectively.}
\end{table*}

\begin{table*}[t]
\centering
\scalebox{0.6}{
\begin{tabular}{l|l|ccccc|cccc|cccc}
\hline\hline
\multirow{2}{*}{Metric} & \multirow{2}{*}{Model} & \multicolumn{5}{c|}{Within Family} & \multicolumn{4}{c|}{Out of Germanic} & \multicolumn{4}{c}{Into Germanic} \\ 
\cline{3-15} 
& & Germanic & Romance & Slavic & Aryan & Asiatic & Romance & Slavic & Aryan & Asiatic & Romance & Slavic & Aryan & Asiatic \\ \hline
\multirow{3}{*}{ChrF} 
& Baseline & 20.8 & 23.5 & 12.1 & 8.0 & 7.2 & 21.0 & 10.5 & 5.8 & 7.3 & 20.9 & 20.0 & 16.8 & 17.8 \\
& Boost-All & 44.6 &\bf 45.6 &\bf 39.2 &\bf 34.2 &\bf 27.4 &\bf 42.2 &\bf 38.0 &\bf 34.1 &\bf 30.9 &\bf 42.6 &\bf 40.6 &\bf 34.8 &\bf 36.4 \\
& Boost-Germanic &\bf 45.0 & 44.5 & 38.5 & 30.3 & 25.0 & 41.3 & 37.4 & 32.4 & 29.4 & 42.4 & 40.5 & 33.2 & 35.9 \\
\hline
\multirow{3}{*}{Comet} 
& Baseline & 54.3 & 58.5 & 50.4 & 45.9 & 41.6 & 56.7 & 47.6 & 39.5 & 41.3 & 54.9 & 55.4 & 52.8 & 50.0 \\
& Boost-All & 68.2 &\bf 71.6 &\bf 74.0 &\bf 71.2 &\bf 66.2 &\bf 66.3 &\bf 69.2 &\bf 69.5 &\bf 68.2 &\bf 67.3 &\bf 67.3 &\bf 63.4 &\bf 61.1 \\
& Boost-Germanic &\bf 68.5 & 70.8 & 72.8 & 68.0 & 63.3 & 65.8 & 68.3 & 68.7 & 66.8 & 66.9 & 66.9 & 61.8 & 60.1 \\
\hline
\multirow{3}{*}{SacreBLEU} 
& Baseline & 2.8 & 3.7 & 2.0 & 3.7 & 0.8 & 2.7 & 1.6 & 2.5 & 1.1 & 2.6 & 2.2 & 1.6 & 1.7 \\
& Boost-All & 17.5 &\bf 18.7 &\bf 13.5 &\bf 15.9 &\bf 6.5 &\bf 16.4 &\bf 12.8 &\bf 15.3 &\bf 8.6 &\bf 15.1 &\bf 13.8 &\bf 9.7 &\bf 11.3 \\
& Boost-Germanic &\bf 17.8 & 17.4 & 12.7 & 14.3 & 5.3 & 15.3 & 12.2 & 14.1 & 7.6 & 14.8 & 13.6 & 9.1 & 10.9 \\
\hline
\end{tabular}}
\caption{\label{table-more-data-more-direction} Zero-shot performance on the EC30 dataset. Boost-All means fine-tuning using all of the NTREX data, while Boost-Germanic means partially fine-tuning using Germanic languages. The results are in three groups: 1) "Within Family" shows the performance within a specific language group, where both the source and target languages belong. 2) "Out of Germanic" shows the average performance that is translated out of Germanic languages, e.g., from Germanic to Romance. and 3) "Into Germanic" shows the average performance that is translated into Germanic languages, e.g., from Romance to Germanic.}
\end{table*}

\begin{table*}[t]
\centering
\scalebox{0.65}{
\begin{tabular}{c|cccccccc}
\hline
\hline
ChrF & en & da & de & es & fi & fr & it & nl \\
\hline
en &  -  & 55.5/55.1/53.6 & 50.6/51.0/49.0 & 46.2/46.3/44.8 & 44.2/43.3/42.8 & 57.4/56.8/55.3 & 47.9/47.0/46.3 & 46.5/46.8/45.3 \\
da & 57.2/57.1/55.4 &  -  & 28.3/44.3/47.2 & 24.8/39.9/42.3 & 22.4/38.5/41.8 & 30.5/45.6/50.4 & 26.8/39.5/43.3 & 23.4/41.9/44.2 \\
de & 53.4/53.1/52.0 & 28.5/44.9/48.7 &  -  & 24.6/39.3/41.9 & 21.4/36.6/40.5 & 30.2/44.4/49.0 & 26.9/39.1/42.6 & 23.4/41.4/43.7 \\
es & 48.0/47.8/46.8 & 26.5/39.7/43.3 & 25.9/38.3/40.8 &  -  & 20.9/34.8/38.1 & 30.5/44.0/47.4 & 28.3/40.6/43.2 & 22.6/38.2/40.7 \\
fi & 45.9/45.8/45.4 & 26.7/39.0/42.7 & 26.8/36.8/40.6 & 25.2/35.9/38.8 &  -  & 28.4/39.4/44.1 & 26.6/35.2/39.3 & 22.5/36.4/39.6 \\
fr & 56.0/55.3/54.0 & 27.3/43.8/47.4 & 26.8/42.7/45.7 & 25.7/42.4/44.5 & 20.7/37.6/41.2 &  -  & 28.4/43.8/46.3 & 23.0/41.2/43.5 \\
it & 50.2/49.6/48.9 & 26.4/40.8/44.4 & 25.8/39.8/43.0 & 26.5/41.6/43.9 & 21.1/35.1/39.3 & 31.2/46.4/49.8 &  -  & 22.9/39.0/41.9 \\
nl & 48.2/47.9/47.0 & 27.0/40.8/43.6 & 26.7/39.9/42.0 & 24.5/37.3/39.5 & 20.7/34.8/37.8 & 28.4/41.3/44.5 & 26.1/36.8/39.5 &  -  \\

\hline
\hline
\end{tabular}}
\caption{\label{table-appendix-Europarl-weak-ChrF} The detailed ChrF results for the baseline, boosted, and upper bound models on the Europarl-8 dataset are presented, encompassing 14 English-centric directions and 42 zero-shot directions.} 
\end{table*}

\begin{table*}[t]
\centering
\scalebox{0.65}{
\begin{tabular}{c|cccccccc}
\hline
\hline
COMET & en & da & de & es & fi & fr & it & nl \\
\hline
en &  -  & 79.5/78.8/76.9 & 72.8/72.5/70.4 & 75.1/74.9/71.6 & 77.5/76.6/75.7 & 75.3/73.8/72.0 & 76.1/74.9/72.8 & 75.4/75.0/73.0 \\
da & 78.6/78.7/77.6 &  -  & 49.7/64.8/71.8 & 55.3/65.3/71.4 & 50.4/68.5/75.6 & 52.7/61.6/70.3 & 53.1/64.6/72.4 & 52.4/68.1/73.5 \\
de & 76.6/76.6/75.8 & 56.1/71.6/76.1 &  -  & 54.2/63.6/69.7 & 49.0/66.2/73.9 & 51.4/61.7/68.6 & 52.6/64.2/71.4 & 51.0/68.5/73.7 \\
es & 75.5/75.5/74.6 & 54.8/66.6/73.7 & 46.9/60.7/67.3 &  -  & 48.2/66.5/73.4 & 53.4/65.5/72.1 & 54.5/69.6/75.8 & 50.3/64.6/70.7 \\
fi & 75.2/74.3/75.0 & 55.6/66.5/73.6 & 47.9/59.3/67.4 & 54.5/62.5/69.2 &  -  & 51.7/59.5/68.6 & 52.7/63.0/69.9 & 51.7/63.1/69.9 \\
fr & 79.1/79.1/77.6 & 57.4/69.6/75.5 & 48.6/63.7/70.6 & 57.9/70.9/75.2 & 50.3/68.9/75.6 &  -  & 56.3/72.5/77.2 & 52.7/68.4/73.2 \\
it & 76.6/76.6/75.8 & 55.8/67.8/74.2 & 47.4/61.3/69.2 & 56.6/70.4/74.8 & 48.8/66.6/74.3 & 54.8/67.3/73.4 &  -  & 50.8/65.6/71.8 \\
nl & 75.7/75.6/75.0 & 55.6/69.2/75.0 & 48.6/64.1/70.2 & 54.0/64.2/69.5 & 48.3/66.0/72.9 & 51.4/61.0/68.4 & 51.7/64.1/70.6 &  -  \\
\hline
\hline
\end{tabular}}
\caption{\label{table-appendix-Europarl-weak-Comet} The detailed COMET results for the baseline, boosted, and upper bound models on the Europarl-8 dataset are presented, encompassing 14 English-centric directions and 42 zero-shot directions.} 
\end{table*}

\begin{table*}[t]
\centering
\scalebox{0.65}{
\begin{tabular}{c|cccccccc}
\hline
\hline
SacreBLEU & en & da & de & es & fi & fr & it & nl \\
\hline
en &  -  & 30.4/30.4/28.1 & 23.0/23.3/21.6 & 18.9/19.2/17.9 & 13.7/12.8/12.9 & 32.6/30.8/30.6 & 19.9/18.9/18.7 & 17.6/18.0/16.9 \\
da & 31.1/31.0/28.3 &  -  & 6.4/16.4/19.5 & 4.2/13.7/15.9 & 3.0/9.7/12.0 & 7.9/18.4/24.3 & 4.9/12.5/15.8 & 3.3/13.4/15.1 \\
de & 26.5/26.5/24.3 & 7.0/18.9/22.8 &  -  & 4.3/13.0/15.7 & 2.8/8.6/11.1 & 7.6/17.6/22.9 & 5.4/12.5/15.1 & 3.5/13.3/15.3 \\
es & 18.4/18.5/17.4 & 5.0/13.0/15.9 & 4.2/10.2/12.1 &  -  & 2.0/6.5/8.2 & 6.9/15.1/20.3 & 5.5/11.8/15.1 & 2.7/9.5/11.4 \\
fi & 18.5/18.4/17.9 & 5.7/13.7/16.8 & 4.8/10.1/13.1 & 4.5/10.5/12.7 &  -  & 6.1/13.5/18.0 & 4.5/9.2/12.7 & 2.8/9.0/11.3 \\
fr & 28.8/28.4/26.2 & 5.8/17.3/20.6 & 5.4/14.2/17.5 & 4.6/15.8/17.8 & 2.4/9.3/11.5 &  -  & 6.1/16.0/18.6 & 3.2/12.5/14.5 \\
it & 20.7/20.6/19.4 & 5.1/13.8/16.7 & 4.5/11.4/14.3 & 5.0/14.2/17.0 & 2.3/7.1/9.5 & 7.9/18.4/22.8 &  -  & 2.9/10.0/12.1 \\
nl & 20.7/20.5/19.2 & 5.6/14.6/17.0 & 5.1/12.2/13.7 & 4.2/11.4/13.3 & 2.4/7.4/8.9 & 6.6/14.7/18.0 & 4.3/10.3/12.1 &  -  \\
\hline
\hline
\end{tabular}}
\caption{\label{table-appendix-Europarl-weak-SacreBLEU} The detailed SacreBLEU results for the baseline, boosted, and upper bound models on the Europarl-8 dataset are presented, encompassing 14 English-centric directions and 42 zero-shot directions.} 
\end{table*}

\end{document}